\pdfoutput=1
\documentclass{scrartcl}

\title{What makes a language easy to deep-learn? Deep neural networks and humans similarly benefit from compositional structure}

\author{Lukas Galke\thanks{%
Dept. of Mathematics and Computer Science, University of Southern Denmark, Odense, Denmark\\
LEADS group, Max Planck Institute for Psycholinguistics, Nijmegen, Netherlands\\email: \texttt{galke@imada.sdu.dk}}\\
 Yoav Ram\thanks{%
School of Zoology, Faculty of Life Sciences, Tel Aviv University, Tel Aviv, Israel\\
Sagol School of Neuroscience, Tel Aviv University, Tel Aviv, Israel}\\
 Limor Raviv\thanks{LEADS group, Max Planck Institute for Psycholinguistics, Nijmegen, Netherlands\\
cSCAN, University of Glasgow, Glasgow, UK}
}
\date{\today}

\usepackage[numbers]{natbib}

\usepackage[english]{babel}

\usepackage{amsmath}
\usepackage{amssymb}
\usepackage{graphicx}
\usepackage[colorlinks=true, allcolors=blue]{hyperref}

\usepackage{todonotes}
\usepackage{xspace}
\usepackage{booktabs}

\usepackage{lineno}

\newcommand\ie{i.\,e.\xspace}

\usepackage{bm}

\begin{document}
\maketitle

\begin{abstract}
Deep neural networks drive the success of natural language processing. A fundamental property of language is its compositional structure, allowing humans to systematically produce forms for new meanings. For humans, languages with more compositional and transparent structures are typically easier to learn than those with opaque and irregular structures. However, this learnability advantage has not yet been shown for deep neural networks, limiting their use as models for human language learning. Here, we directly test how neural networks compare to humans in learning and generalizing different languages that vary in their degree of compositional structure. We evaluate the memorization and generalization capabilities of a large language model and recurrent neural networks, and show that both deep neural networks exhibit a learnability advantage for more structured linguistic input: neural networks exposed to more compositional languages show more systematic generalization, greater agreement between different agents, and greater similarity to human learners.

\end{abstract}

Compositionality, i.e., whether the meaning of a compound expression can be
derived solely from the meaning of its constituent parts, has been studied for decades by
both computer scientists and linguists~\cite{DBLP:conf/iclr/Andreas19,lakeHumanlikeSystematicGeneralization2023,sep-compositionality,fodor2002compositionality,janssenFregeContextualityCompositionality2001}.
In particular, languages differ in how they map meanings into morphosyntactic
structures~\cite{dryer_world_2013,evans_myth_2009} and cross-linguistic studies
find substantial differences in the degree of
structural complexity across languages~\cite{ackerman_morphological_2013,bentz_learning_2016,hengeveld_transparent_2018,lewis2016linguistic,lupyan2010language,mccauley_language_2019,wuMorphologicalIrregularityCorrelates2019a}.
These differences can stem from multiple and often confounded aspects of
linguistic structure including the degree of compositionality~\cite{evans_myth_2009},
which can be quantified by correlating
differences in meaning with differences in form~\cite{DBLP:journals/alife/BrightonK06}.
For example, the English term ``white horse'' is compositional since its
meaning can be directly inferred given knowledge about its
constituents ``white'' and ``horse''. In contrast, consider the equivalent
German term ``Schimmel'', whose meaning cannot be derived from ``weiß'' (white) and ``Pferd'' (horse).
Crucially, compositionality directly affects our ability to make systematic generalizations in a given
language and thus shapes its immense expressive power -- which also explains its high relevance in machine
learning~\cite{lakeHumanlikeSystematicGeneralization2023,akyurek-andreas-2023-lexsym,hupkes2023stateoftheart,xu2022compositional,hupkes2020compositionality,scan,cogs,baroni2020linguistic,DBLP:conf/atal/Resnick0FDC20,DBLP:conf/iclr/Andreas19,word2vec}.

Importantly, cross-linguistic differences in compositional structure were suggested to
impact human language learning and generalization in the real world~\cite{dekeyser_what_2005,kempe_second_2008,kempe_acquisition_1998} as
well as in lab experiments~\cite{raviv2021easy2learn,kirby_cumulative_2021,raviv_systematicity_2018,cornish_sequence_2017,kirby_cumulative_2008}, with more compositional linguistic
structures typically being easier to learn for adult learners.
In a large-scale artificial language learning study with adult human participants, the
acquisition of a broad yet tightly controlled range of comparable languages
with different degrees of compositional structure was tested~\cite{raviv2021easy2learn}.
Results showed that more compositional languages were learned faster, better, and more
consistently by the adult learners, and that learning more structured languages
also promoted better generalizations and more robust convergence on labels for
new, unfamiliar meanings. This is likely because more systematic and
compositional linguistic input allow learners to derive a set of generative
rules rather than rote memorizing individual forms, and then enables learners
to use these rules to produce an infinite number of utterances after exposure
to just a finite
set~\cite{kirby_learning_2002,kirby_ug_2004,zuidema2002poverty,kirby_cumulative_2008,tamariz_cultural_2016}.
This learnability and generalization advantage for more structured linguistic
input has far-reaching implications for broader theories on language evolution
in our species (and potentially other learning systems): A large body
of computational models and experimental work with human participants show that
more systematic and compositional structures emerge during cross-generational
transmission and communication precisely because such structures are
learned better, while still allowing for high
expressivity~\cite{kirby_learning_2002,kirby_ug_2004,zuidema2002poverty,kirby_cumulative_2008,kirby2015compression,raviv_systematicity_2018,motamedi_evolving_2019,motamedi_emergence_2021,carr_simplicity_2020}.
Hence, popular theories of language evolution attribute the emergence of systematic and
compositional structure in natural languages to such learnability
pressures~\cite{kirby_cumulative_2008,tomasello2005constructing}, suggesting
a causal role not only in language learning, but also in shaping the way human languages are structured.
To what extent this advantage of linguistic structure carries over to
artificial learning systems is currently poorly understood -- which is the aim
of the current study.

Despite an increasing body of work that reports striking similarities between
humans and large language
models~\cite{liImplicitRepresentationsMeaning2021,patelMappingLanguageModels2022,li2022emergent,abdouCanLanguageModels2021,srikant2022convergent,schrimpfNeuralArchitectureLanguage2021,dasguptaLanguageModelsShow2022},
and despite large language models being incredibly proficient at using language
and generalizing to new tasks with little to no new training
data~\cite{gpt3,weiEmergentAbilitiesLarge2022,bommasaniOpportunitiesRisksFoundation2022,radford2019language},
research on emergent communication suggests that deep neural networks (the
class of models that underlies large language models) show no correlation
between the degree of compositional structure in the emergent language and the
generalization capabilities of the networks. In other words, unlike humans,
artificial neural networks do not seem to benefit from more
compositional structure when they are made to develop their own communication
protocol, at least without dedicated
intervention~\cite{rita_emergent_2022,DBLP:conf/acl/ChaabouniKBDB20,kottur-etal-2017-natural,DBLP:conf/nips/LiB19} (but see~\cite{conklin2023compositionality}).
Thus, this finding raises the question of whether systematic and compositional
linguistic structure is helpful at all for deep neural networks, and to what
extent compositionality affects the memorization and generalization
abilities of deep neural networks learning a new language.

The mismatch with humans can potentially be explained by differences in model design and
experimental procedure~\cite{galke2022emergent}. For instance, deep neural
networks typically have immense model capacity due to
overparametrization~\cite{nakkiran2021deep,kaplanScalingLawsNeural2020,belkin2019reconciling,pmlr-v80-arora18a,mackay2003information,DBLP:journals/mcss/Cybenko89},
which means they could easily memorize all individual forms without the need to
identify compositional patterns~\cite{DBLP:conf/atal/Resnick0FDC20,galke2022emergent}. A competing
hypothesis is that neural networks do benefit from compositional structure in
the data given that this structure is reflected in the statistical patterns of
the data which impacts the optimization of the model parameters~\cite{carlini2022memorization,tirumala2022memorization}.
Specifically, in a language with a higher degree of compositionality, the
individual units of meaning are reused in different contexts and thus appear
more often in the training data, such that these recurring units of meaning and
their contextualization patterns are learned better because of the
repeated presentation throughout training
(cf.~\cite{harris_distributional_1954,word2vec}).

\begin{figure*}
    \centering
    \includegraphics[width=0.99\textwidth]{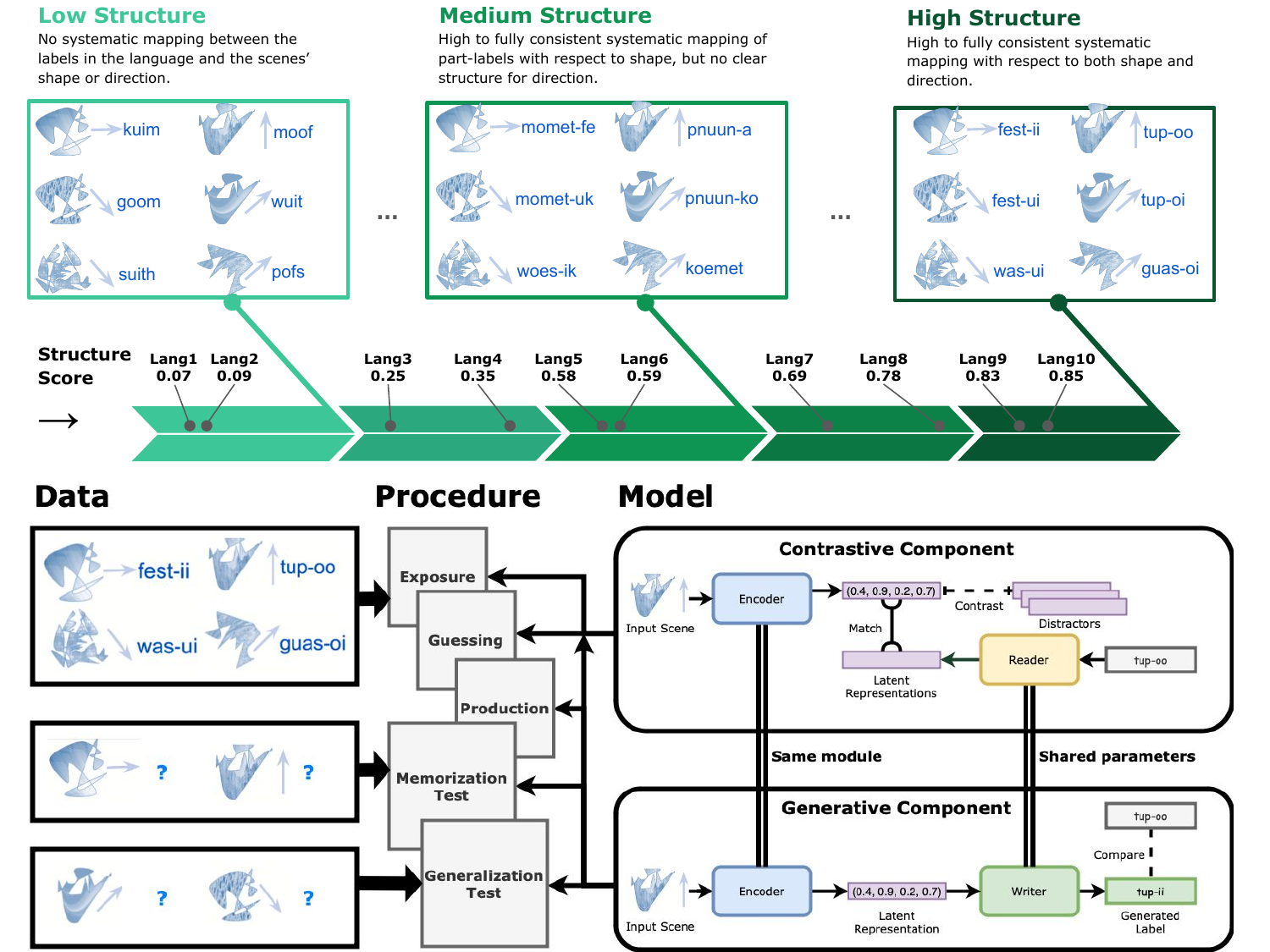}
    \caption{Overview of input languages (Top), the experimental procedure (Bottom Center) along with exemplary input data from one language (Bottom Left), and the model architecture (Bottom Right).
Low-structured input languages show no signs of systematicity or
compositionality, whereas high-structured languages are systematic and compositional with respect to both
attributes: shape and angle. For each language, we train the model for multiple rounds of
exposure, guessing, production. After each round, we conduct a memorization test to evaluate productions for previously
seen items and a generalization test evaluating the productions for new items. Graphical elements in the upper part of this figure are re-used and adapted with permission from \citet{raviv2019larger}.}\label{fig:one}
\end{figure*}

Here, we explore this precise relationship between compositional structure and
generalization with deep neural networks.  The central question we aim to
answer is: Do deep neural network models exhibit the same learning and
generalization advantage when trained on more structured linguistic input as
human adults?  Specifically, we investigate whether the advantage of
compositionality in language learning and language use carries over to
artificial learning systems, while considering GPT-3.5 as a pre-trained
large language model and a custom model architecture based on recurrent neural
networks (RNNs) trained from scratch.  Our work contributes to the
understanding of deep neural networks and large language models, sheds new
light on the similarity between humans and machines, and, consequently, opens
up future directions of simulating the very emergence of language and
linguistic structure with deep neural network agents.

To allow for direct comparisons between humans and
machines, we carefully follow the experimental procedure and measures of a
recent large-scale preregistered language learning study with adult
participants~\cite{raviv2021easy2learn}. We consider 10 input languages, each of which has emerged independently and spontaneously through a group communication experiment with adult human participants~\cite{raviv2019larger}. The languages describe four different novel shapes moving on the screen in a different direction (0-360 degree), and vary in their degree of compositional structure: ranging from fully idiosyncratic languages with entirely different labels for two related meaning (e.g., 'kuim' and 'goom' for the same shape moving into a different direction) to highly structured languages, which re-use parts of the descriptive label (e.g., referring to the two scenes as 'fest-ii' and 'fest-ui'). See Figure~\ref{fig:one}.
Neural networks were then trained on the
exact same stimuli presented to humans and in the same order, using the same
learning tasks, providing the same feedback during learning blocks, and
evaluated with the same memorization and generalization tests.
Figure~\ref{fig:one} shows the recurrent neural network architecture and summarizes the
experimental procedure:
Full details of the experimental setup, custom recurrent neural neural
network models, and how we employed large language models are provided in the Methods section.

\section*{Results}\label{sec:results}

To preview our results, we find a consistent advantage of more systematic and compositional linguistic
structure for learning and generalization, closely reflecting adult human
participants. The generalization behavior of both large language models (pre-trained on other
languages) and recurrent neural networks (trained from scratch) was far more
systematic and transparent when the input languages were more compositional.
Moreover, recurrent neural network agents displayed a higher agreement with
other agents as well as with humans when the input was more compositional,
leading to converging transparent generalizations for new unseen input. A glossary of evaluation metrics can be found in Table~\ref{tab:glossary}. More detailed descriptions of the metrics are provided in the Methods section.

\begin{table}[]
    \centering
    \caption{Glossary of Metrics}
\begin{tabular}{lp{10cm}}
\toprule
    \textbf{Metric} & \textbf{Description} \\
    \midrule
     Production Similarity & One minus length-normalized edit distance\\
     Semantic Difference & Sum of the difference in shape (1 if different and 0 otherwise) and the absolute difference in angles (divided by 180)\\
     Structure Score  & Pearson correlation between (a) pairwise semantic distances and (b) pairwise normalized Levenshtein distances, where (a) and (b) are calculated on all pairs of items in the original input language \\
     Generalization Score & Pearson correlation between (a) pairwise semantic difference and (b)~pairwise length-normalized edit distance, where (a) and (b) are calculated on all pairs between productions for memorized items and productions for generalized items\\
     Convergence Score & Average of all values for item-level production similarity for the same items between different learners trained on the same language\\
     Human Label Similarity &  Item-level production similarity to (other) human learners, averaged across different human learners\\
     True Label Similarity &  Item-level production similarity to input language\\
     \bottomrule
    \label{tab:glossary}
    \end{tabular}
\end{table}

\subsection*{More compositional structure leads to higher similarity to humans and more systematic generalization of large language models}
\begin{figure*}[ht!]
    \centering
    \includegraphics[width=0.92\linewidth]{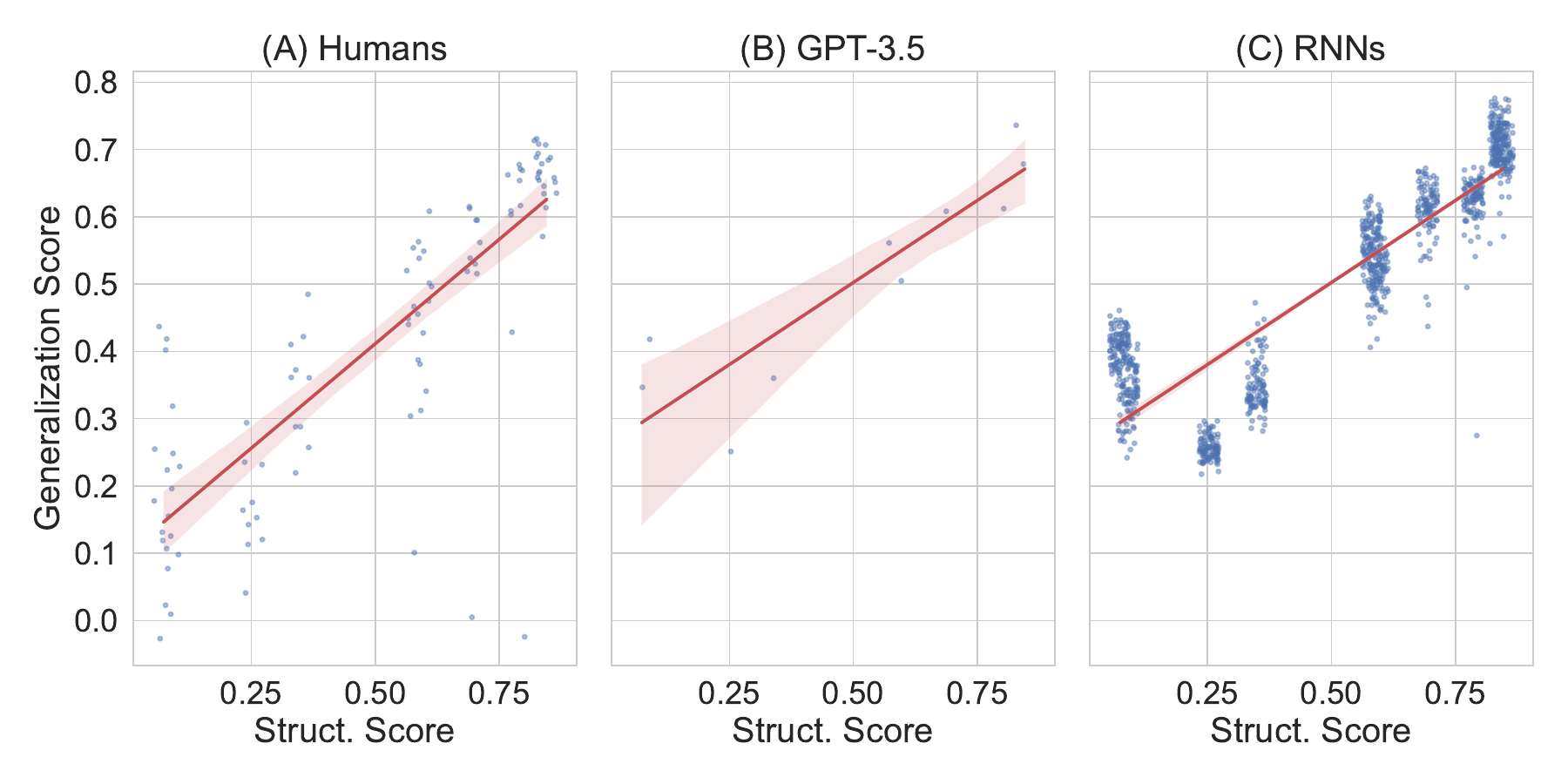}
    \caption{\textbf{Final Generalization Score} achieved by humans (A), GPT-3.5 (B), and recurrent neural networks (C) for each of the input languages. The x-axis shows the structure score of the input languages. Each point corresponds to  the generalization score calculated for the entire input language. This score reflects the degree to which learners systematically generalized new labels relative to the labels they learned. For example, generalization score would be high if learners successfully recombines previously used parts, e.g., combining 'muif' for the shape and 'i' for the direction into 'muif-i'.  Error regions of the regression lines are 95\% confidence intervals estimated via bootstrapping. More structure in the input language leads to more systematic generalization for all three learning systems.}\label{fig:gpt3:generalization}
\end{figure*}

We first test whether large language models benefit from
compositional structure when learning a new language. Such language models are
pre-trained to predict left-out words in web-scale corpora of text data,
leaving them with high competence in at least one language, similar to adult human participants.
Specifically, we employ the large language model GPT-3.5 (version
text-davinci-003) which is capable of in-context learning, \ie, having the
model tackle a new task only based on a few examples in the prompt~\cite{gpt3,rlhf}. We make use of this property to evaluate the model in learning the new languages.
For each input language, we insert the form-meaning pairs in the prompt of
the large language model, followed by a single meaning for which the label
needs to be completed.  We repeat this procedure multiple times to have the language model produce labels for the
memorization test (known meanings) as well as the generalization test (new
meanings).

\begin{table}[ht]
    \centering
    \caption{Generalization examples from neural network and human learners, showing labels generated for unseen scenes. The column GPT-3.5 corresponds to completions generated by the GPT-3.5 model text-davinci-003 via in-context learning, where the training data is provided in context. The examples cover the differently structured input languages from low to high.}
    \label{tab:reg-examples}
    \begin{tabular}{lrrlll}
\toprule
\textbf{Struct.} &  \textbf{Shape} &  \textbf{Angle} & \textbf{Human} & \textbf{RNN} & \textbf{GPT-3.5} \\
\midrule
   low &      2 &    360 &        kokoke &      seefe & tik-tik\\
    &      4 &     45 &          woti &       kite & hihi\\
    &      3 &    150 &          ptiu &       mimi & hihi\\
   \midrule
   mid-low &      3 &    225 &      wangsuus &    wangsoe & wangsuus\\
    &      4 &    225 &        gntsoe &      gntuu & gntsii\\
    &      1 &    135 &        sketsi &       gesh & geshts\\
   \midrule
   mid &      3 &     60 &          powi &   powu-u-u & powee\\
    &      4 &    330 &       fuottoa &     fuotio & fuottu-u-u \\
    &      1 &     30 &    fewo-o-o-o &      fewen & fewee\\
   \midrule
   mid-high &      1 &     30 &         fas-a &      fas-a & fas-a \\
    &      3 &    360 &        muif-i &     muif-a & muif-i \\
    &      1 &    225 &      fas-huif &   fas-huif & fas-huif\\
   \midrule
   high & 4 & 60 & smut-tkk & smut-tk & smut-ttk \\
    & 2 & 360 & nif-k & nif-kks & nif-k \\
    & 1 & 315 & wef-ks & wef-kks & wef-kks \\
\bottomrule
\end{tabular}
\end{table}

In the generalization test, there is no true label in the input language. To
capture the degree to which new labels conform to the labels of the input
language (i.e., to what extent the generalization is systematic), we correlate the pairwise label difference and the pairwise semantic difference between the labels generated for new scenes and the labels generated by the same agent for known
scenes~\cite{raviv2021easy2learn}.

Strikingly, the results reveal that a higher degree of compositional structure
in the input language leads to generalizations that are more systematic (see
Figure~\ref{fig:gpt3:generalization}B), closely reflecting the pattern of
adult human learners (Figure~\ref{fig:gpt3:generalization}A).
Table~\ref{tab:reg-examples} shows examples of the final productions of humans
and large language models during generalization (more examples are provided in Tab.~6 and 7 in the SI).

\begin{figure*}[ht!]
    \centering
    \includegraphics[width=0.92\linewidth]{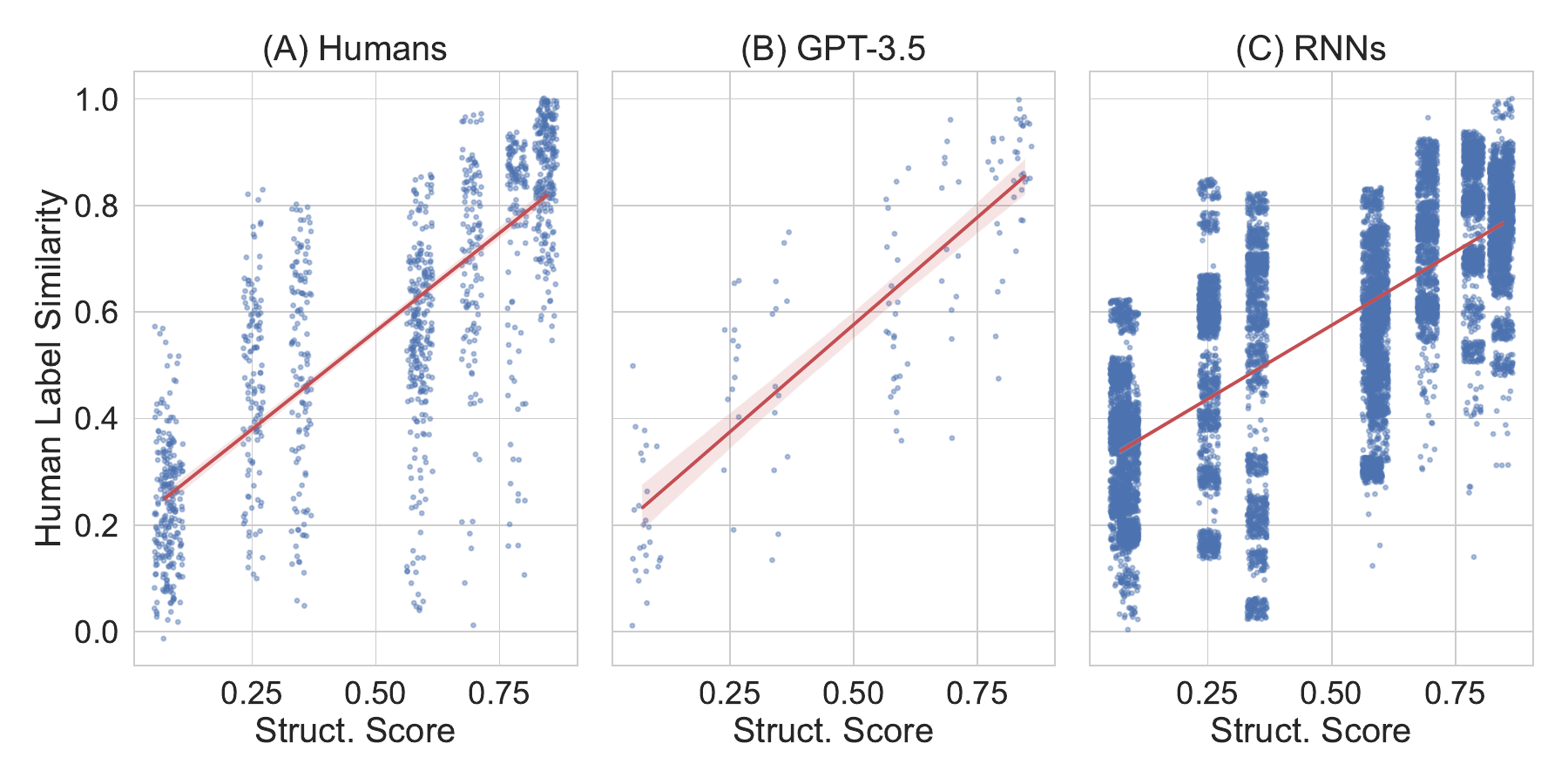}
    \caption{\textbf{Final Similarity to Humans during Generalization} Final production similarity with (other) human participants during
generalization achieved by humans (A), GPT-3.5 (B) and recurrent neural networks (C) for each of the input languages. The x-axis shows the structure
score of the input languages. Each point corresponds to the production similarity score (calculated as length-normalized edit distance) between humans' productions and models' productions for every  item in the language. For example, a recurrent neural network that produced 'muif-a' for shape 3 moving in direction 360 degrees would have a high production similarity to the majority of human participants who produced 'muif-i'.
Error regions of the regression lines show 95\% confidence intervals estimated via bootstrapping. More structure in the input language leads to more similarity to human participants for both RNNs and GPT-3.5.}\label{fig:gpt3:sim2humans}
\end{figure*}

In addition, we evaluate the production similarity as character-level length-normalized edit distance between the generated labels and labels produced by human participants during generalization. The results show that, given more
structured linguistic input, GPT-3.5 also yields productions that are more
similar to the productions of human participants, calculated as the average
similarity between GPT-3.5's production and all human productions for the same scene in the same language
(Figure~\ref{fig:gpt3:sim2humans}B).
Analogously, Figure~\ref{fig:gpt3:sim2humans}A shows the similarity of humans to other human learners during generalization.

\begin{figure}[ht!]
    \centering
    \includegraphics[width=0.92\linewidth]{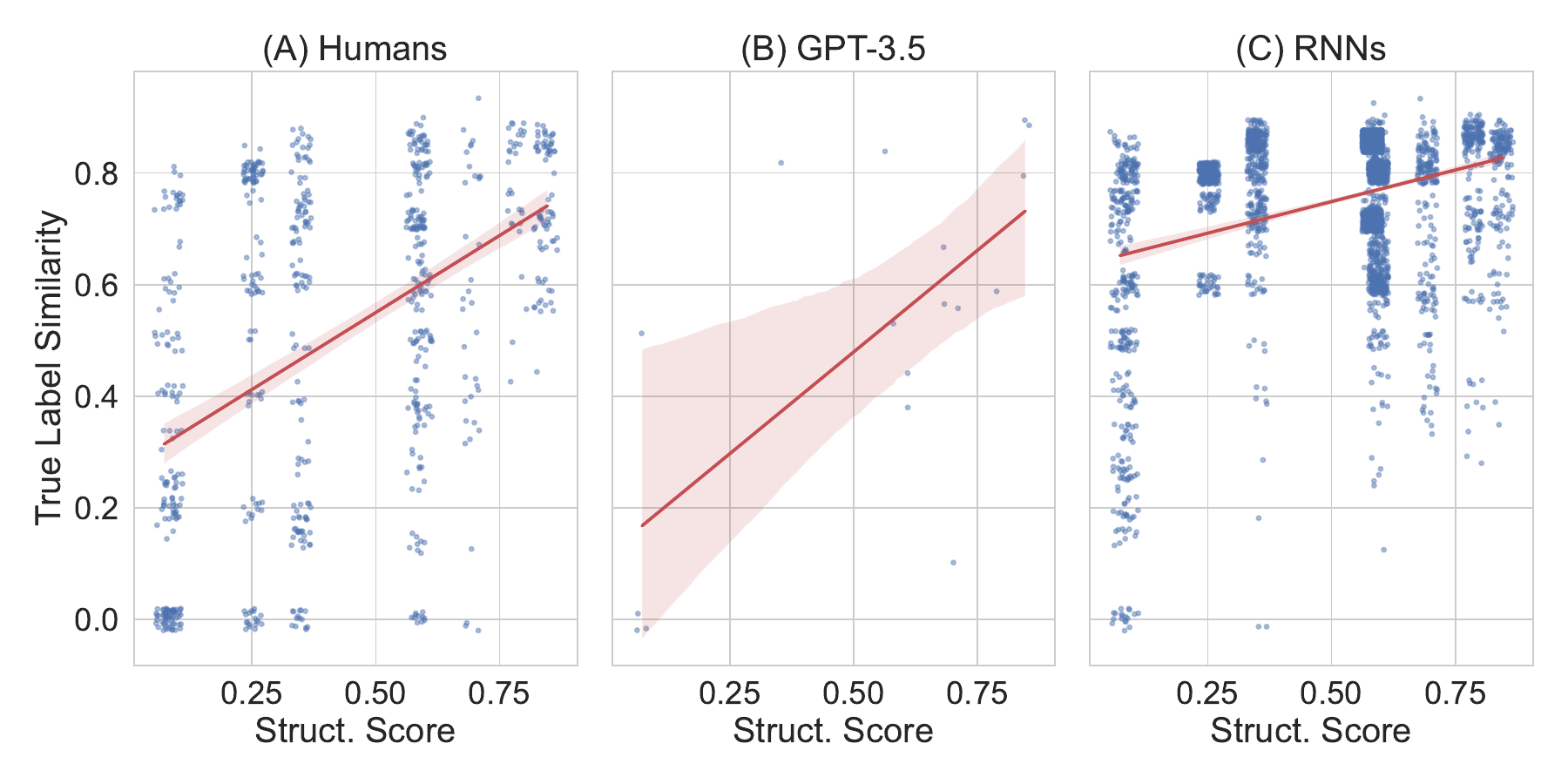}
    \caption{\textbf{Memorization Error Analysis} for human participants (A), GPT-3.5 (B), and recurrent neural networks (C). The error rates are 33.30\% for humans, 7.39\% for GPT-3.5 via in-context learning, and 13.87\% for RNNs after 100 epochs of training. The x-axis shows the structure score of the input language. Each point corresponds to the production similarity score (calculated as length-normalized edit distance) between an erroneously memorized label for a given item and the correct corresponding label as it appears in the input language. For example, 'wangsus' has a higher similarity with 'wangsuus' than 'gempt'.
    Error bands of the regression lines show 95\% confidence intervals estimated via bootstrapping. More structure leads to erroneously memorized examples being more similar to the ground truth of the input language.}\label{fig:gpt3:error-analysis}
\end{figure}

We then conduct an error analysis to understand better whether the memorization
errors are similarly affected by the degree of compositional structure.
We analyze the cases where the learning system fails to memorize the correct
label perfectly and calculate the production similarity (1 minus
length-normalized edit distance). Again, the results show
the same pattern for adult human participants and large language models
(see Figure~\ref{fig:gpt3:error-analysis}A and B): When there is more structure in the input
language, the non-perfectly memorized productions are more similar to the
correct labels.

\subsection*{More compositional structure leads to higher similarity to humans and more systematic generalization with recurrent neural networks}

\begin{figure*}[htp]
\centering
\includegraphics[width=\textwidth]{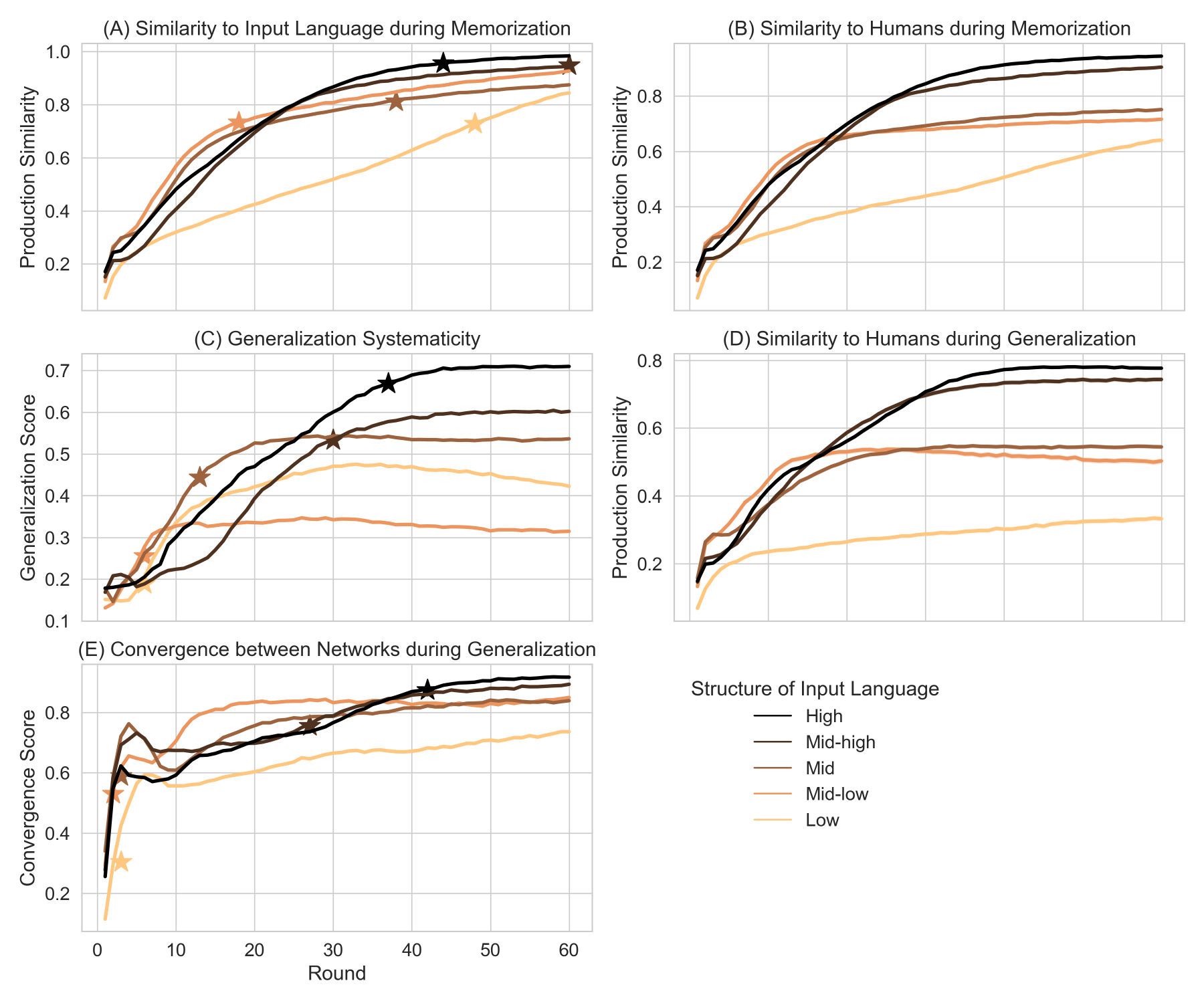}
\caption{
\textbf{Learning trajectory of Recurrent Neural Networks' Memorization and Generalization Performance}. More structured languages lead to better and faster reproduction of the input language (A), to better generalization on unknown scenes (C), better agreement with human participants during memorization (B) and generalization (D), and higher convergence between networks (E).
(A): Production similarity between labels generated by neural agents and labels of the input language.
(B): Production similarity between labels generated by neural agents and labels generated by human participants.
(C): Generalization score of labels generated by neural agents for new scenes that were not part of the training data.
(D): Production similarity between labels generated by neural agents and labels generated by human participants for unseen scenes.
(E): Convergence score measures the similarity between labels generated for unseen scenes by different neural agents.
Stars mark the round at which neural agents first exceed the final performance of human participants.
Input languages are grouped into 5 bins. Each line is the average of 200 neural
agents with a different random initialization. A star marks the epoch at which
the RNN agents exceed human performance. Results are cut off for visualization
at epoch 60, full results in SI.
}\label{fig:big-panel:60}
\end{figure*}

In addition to large language models, we test a custom neural
network architecture trained from random initialization, which allows us to conduct a close analysis of the learning trajectory.
Our custom model architecture is designed to simulate the exposure, guessing, and production that human participants were exposed to (see
Figure~\ref{fig:one}).  The architecture is inspired by image-captioning
approaches~\cite{vinyals_show_2015}, the emergent communication
literature~\cite{lazaridou2020emergent}, and in particular, our recent review
paper~\cite{galke2022emergent} which suggested having shared model parameters
between generation and processing of a label. Our model consists of two
components: a generative component that facilitates the production of a
descriptive sequence of symbols (here, a label) for a scene, while a
contrastive component shapes the latent space and enables the models to carry
out guessing tasks during learning (i.e., given a label, pick the correct scene
from a set of distractors). Each component has a sequential recurrent neural
network module to carry out the generation and processing of a sequence,
respectively, for which we use the well-known long short-term
memory~\cite{lstm}. The symbol embedding that maps each symbol of the sequence
into a continuous vector is shared between the generative and the contrastive
component. Moreover, the two components share the same encoder module that
transforms an input scene into a latent representation, which then serve as the initial
state of the generative component. Production tasks are modeled by a
generative objective: Based on this initial state, the model generates a label,
character by character. This generated label is compared to the target input
language by character-level cross-entropy. Guessing tasks are modeled by a
contrastive objective~\cite{simclr}, which aligns the latent representation of
input scenes and corresponding labels and facilitates selecting the correct
scene from a set of distractors. As the encoder is shared, the contrastive
objective shapes the space of initial states of the production model.

In total, we trained 1,000 neural network agents with different random seeds
(100 for each of the ten input languages) and calculated the following measures
after each training round: the similarity between networks' productions and the
input language; the similarity between networks' productions and the human
learners' productions during memorization and generalization; the
generalization score capturing the degree of systematicity; and a convergence
score capturing the agreement between different agents. We evaluated these
measures after each of 100 rounds.

The results are shown in Figure~\ref{fig:big-panel:60}. Extended results can be
found in Figs.~6--11 in the SI. In the following, we present the results for
the learning trajectory organized along the two types of tests: memorization
and generalization, before presenting the final results of RNNs.

\paragraph{Memorization Trajectory}
How well did neural agents memorize the input
languages? And how similar were their generated labels to those produced by
human learners during generalization? This is measured by production
similarity~\cite{raviv2021easy2learn}, which captures the similarity between
the original label and the produced label by calculating the average normalized
edit distance between two labels for the same scene. We use this measure in
two ways: once to compare the generated labels to the true label of the input
language and once to compare the machine-generated label to the human-generated
label for the same scene.

\paragraph{Similarity to Input Languages during Memorization} With sufficient
training rounds, all languages can be learned by all neural network agents,
reaching a production similarity of at least 0.8 (out of 1) by round 60
(Figure~\ref{fig:big-panel:60}A). Structured languages are learned significantly
better (LME~1\xspace; $\beta = 0.045$, $\mathrm{SD}=0.001$, $z=62.865$,
$p<0.001$), \ie, they show a higher similarity with the input language.
However, this advantage tends to diminish over training rounds
(LME~1\xspace; $\beta = -0.005$, $\mathrm{SD}<0.001$, $z=-54.978$,
$p<0.001$).

\paragraph{Similarity to Humans during Memorization} We measure the similarity to humans during memorization (\ie, comparing
productions of both learning systems after completing the training rounds) and
the memorization test data of the neural network agents after each training
round (Figure~\ref{fig:big-panel:60}B). More compositional input languages led to a
significantly greater similarity with human learners (LME~2\xspace;
$\beta = 0.097$, $\mathrm{SD}=0.001$, $z=81.429$, $p<0.001$). This effect
became even stronger over rounds (LME~2\xspace; $\beta = 0.022$,
$\mathrm{SD}=0.000$, $z=208.708$, $p<0.001$).

\paragraph{Generalization Trajectory}
We evaluate the productions of neural agents when they generalize, \ie, produce
labels for new scenes that were not part of the training data. We test the
productions regarding three aspects: the degree of systematicity, the
similarity to humans, and the generalization convergence between different
agents.
As with large language models, we evaluate the generalization
score. More structured languages consistently led to significantly higher
generalization scores (Figure~\ref{fig:big-panel:60}C) (LME~3\xspace; $\beta =
0.088$, $\mathrm{SD}=0.001$, $z=148.901$, $p<0.001$), and this effect became
stronger with time ($\beta = 0.046$, $\mathrm{SD}<0.001$, $z=703.483$,
$p<0.001$).

\paragraph{Similarity to Humans during Generalization} We measure the similarity
between the productions of neural network agents and humans for new scenes
(Figure~\ref{fig:big-panel:60}D), \ie, during generalization. Examples are shown in Table~\ref{tab:reg-examples}.  More structure in
the input language led to a significantly higher similarity between humans and
neural agents (LME~5\xspace; $\beta = 0.132$, $\mathrm{SD}=0.002$,
$z=70.280$, $p<0.001$), which became stronger over rounds ($\beta = 0.046$,
$\mathrm{SD} < 0.001$, $z=344.287$, $p<0.001$).

\paragraph{Convergence between Neural Agents during Generalization} More structured
languages lead to better agreement between networks (LME~4\xspace; $\beta
= 0.043$, $\mathrm{SD}=0.001$, $z=49.027$, $p<0.001$), such that, for more
structured languages, different neural agents learning the same input language
produced more similar labels for new scenes (Figure~\ref{fig:big-panel:60}E). This
effect became stronger over rounds ($\beta = 0.009$, $\mathrm{SD}<0.001$,
$z=121.740$, $p<0.001$).

\paragraph{Final Results of RNNs}
To compare our custom recurrent neural network agents with large language
models and with humans, we visualize the relationship between compositional
structure of the input language and final generalization performance in
Figure~\ref{fig:gpt3:generalization}C. All three learning systems (Humans, RNNs, and
GPT-3.5) show the same trend: more compositionality in the input language leads
to more systematic generalization.

Moreover, we calculate the average similarity to generalizations of human participants on the same language and item. Comparing the
productions during generalization, the results show
that a higher degree of structure in the input language leads to more
similarity with humans (see Figure~\ref{fig:gpt3:sim2humans}C). This pattern
of compositional structure leading to more human-like generalizations
is present in both RNNs' and GPT-3.5's generated labels -- as well as when comparing humans to other humans (see Figure~\ref{fig:gpt3:sim2humans}).

Lastly, we visualize the results of the memorization
error analysis for recurrent neural networks alongside humans and GPT-3.5 in
Figure~\ref{fig:gpt3:error-analysis}. The pattern is the same for all three
different learning systems, be it artificial or biological: more compositional
structure leads to errors that are more similar to the true label.

\section*{Discussion}\label{sec:discussion}

Our results show that deep neural networks benefit from
more structured linguistic input as humans do and that neural networks'
performance becomes increasingly more human-like when trained on more
structured languages. This structure bias can be found in networks' learning
trajectories and even more so in the networks' generalization behavior,
mimicking previous findings with humans. Although all languages can eventually
be (almost) perfectly learned, we show that more structured languages are learned better and more
similarly to human productions. Deep neural networks and humans produce nearly identical
labels when trained on high-structured languages but not when trained on
low-structured languages.  Moreover, networks that learn more structured
languages are significantly better at systematic generalization to new, unseen
items, and crucially, their generalizations are significantly more consistent
and more human-like. This means that highly systematic grammars allow for
better generalization and facilitate greater alignment between different neural
agents and between neural agents and humans. We have replicated these results with
small recurrent neural networks and with transformer-based large language
models, showing that, together with humans, all three learning systems show the
same bias in systematic generalization and memorization errors. Thus,
our findings strengthen the idea that language models are useful for studying
human cognitive mechanisms, complementing the increasing evidence of similarity
in language learning between humans and
machines~\cite{liImplicitRepresentationsMeaning2021,patelMappingLanguageModels2022,li2022emergent,abdouCanLanguageModels2021,srikant2022convergent,schrimpfNeuralArchitectureLanguage2021,dasguptaLanguageModelsShow2022}.

Specifically, we find very similar effects of  structure on generalization and on the similarity to humans across all three learning systems. While we find a different slope for humans and RNNs in the memorization error analysis (likely due to RNNs being less impacted by memorization difficulty given sufficient training),  the general trend is consistent: for both humans and artificial agents, exposure to more structured languages leads to production errors that are nevertheless more similar to the correct labels (i.e., their errors are less ``wrong'').

We assume that the reason for the increased similarity between machines and humans is that the ways to generalize are more transparent in high-structured languages, while there are none or less transparent generalization patterns available in low- and medium-structured languages. This leads both humans and neural networks to a higher production variation in lower structured languages, as different options on how to generalize are equally likely. This point is well supported by results from humans, who indeed show increased convergence between participants when learning higher structured languages~\cite{raviv2021easy2learn} Our results thereby demonstrate that what is more transparent for humans is also more transparent for deep neural networks.

Analyzing the learning trajectory of recurrent neural networks, we find that languages with mid and mid-low structures often show an advantage in both memorization and generalization during the early stages of learning. This may be due to the fact that these mid-structured languages trade off full expressiveness with more simplicity (see Tab. 3 in the SI). For example, one of the mid-structured languages includes a marker  for ``moving on the diagonal'', but does not distinguish the direction of the movement (e.g., center to north-east vs.\@ center to south-west). As a result, the same label is used for two distinct meanings, which is easier to learn in the first place (less variation), but not sufficient to fully differentiate between items and thus harming systematic generalization.

As for implications, our findings first and foremost support the idea that languages' underlying
grammatical structure can be learned directly from (grounded) linguistic input
alone~\cite{vongGroundedLanguageAcquisition2024,piantadosiModernLanguageModels2023,piantadosi_infinitely_2017,tomasello2005constructing,zuidema2002poverty}.
To ensure that the advantage of more structured linguistic
input does not  stem from the fact that the learning system was already proficient
in a different language -- \ie, as are pre-trained language models and adult humans
-- we also also considered models trained from random initialization.
Therefore, our results predict that children would also benefit
from more systematic compositional structure in the same way adults do -- a
prediction we are currently testing (preregistration:
\cite{lammertinkLearnabilityEffectsChildren2022}).

Our findings have further implications for machine learning, where
systematic generalization beyond the training distribution (out-of-domain) is of high interest~\cite{hupkes2023stateoftheart,hupkes2020compositionality,diera-etal-2023-gencodesearchnet,scan,cogs}.
Systematic in-domain generalization, as studied here, is a critical prerequisite for systematic out-of-domain generalization.
Specifically, we show that seeding a learning system with well-structured inputs can
improve their ability to systematically generalize to combinations that were not observed during training.  Even though
our study is based on artificial languages, our findings directly pertain to
the natural language processing of real-world languages. To confirm this
prediction, we re-analyzed data from
Wu~et~al.~\cite{wuMorphologicalIrregularityCorrelates2019a}, who used the Wug
Test~\cite{wugtest} to test language models' ability to predict different forms
of unfamiliar words in a wide range of natural languages. Indeed, we find that
the Wug Test accuracy negatively correlates with the degree of irregularity of
the language (Spearman's $\rho = -0.96$, $p<10^{-15}$; Kendall's $\tau =-0.86,
p< 10^{-14}$). This strong negative correlation suggests that natural languages
with fewer irregularities, \ie, more consistently structured natural languages, are indeed easier to learn for machines.

Crucially, there is a positive correlation between the degree of linguistic
structure and population size~\cite{lupyan2010language,meir_influence_2012,bentz_adaptive_2015,raviv2019larger},
with low-resource languages (\ie, languages spoken by smaller communities for
which there is only very little training data available) typically having less
structured languages. Since our study predicts that such languages are harder
to learn for deep neural networks, this results in a double whammy for
developing natural language processing systems for small communities' languages
-- exacerbating challenges of low-resource language modeling~\cite{aji-etal-2022-one}. Interestingly, the benefit of structured input could also explain the importance of highly-structured programming languages in the data mix for training large language models~\cite{aryabumi2024codecodeexploringimpact}.

Finally, our results are of high relevance to the field of emergent communication. Emergent communication strives to simulate the evolution of language with multi-agent reinforcement learning~\cite{DBLP:conf/iclr/LazaridouPB17,DBLP:conf/iclr/LazaridouHTC18,DBLP:conf/nips/LiB19,lazaridou2020emergent,DBLP:conf/iclr/RenGLCK20,DBLP:conf/nips/MuG21,rita_emergent_2022,rita2022on,chaabouni2022emergent}.
However, as argued in the introduction, certain linguistic phenomena of natural language appear to be hard to replicate in multi-agent reinforcement learning~\cite{kottur-etal-2017-natural,chaabouni2019anti,DBLP:conf/acl/ChaabouniKBDB20,galke2022emergent}, which had raised the question whether compositionality is helpful for neural networks at all.
We hypothesized that these mismatches are caused by the lack of cognitive constraints~\cite{galke2022emergent} eradicating the learnability pressure underlying human language evolution~\cite{kirby2015compression}.
Our findings support the importance of a learnability pressure for compositional languages to emerge.
By confirming a result previously found in humans~\cite{raviv2021easy2learn}
in deep neural networks, we take the first steps to bring emergent
communication closer to the field of language evolution, supporting simulations of language emergence with neural networks.

An interesting direction for future research is to investigate potential differences in the amount of training that a neural network needs compared to humans. Through anchoring our experiments in human data, we were able to directly identify the point during training at which recurrent neural networks equalize with human participants. However, the location of this point depends on various factors such as the amount of data, the number of parameters that are optimized, and the number of optimization steps -- which makes it challenging to predict this point in advance. While we have identified this point through analyzing the learning trajectory, our analysis does not depend on it, as all measures including the similarity between humans and machines are calculated based on productions taken at the end of training.

Furthermore, we have chosen to work with an input representation that we deemed easiest to process for each type of learning system. Since the particular way in which agents represent the visual world was not the object of the current study, our rationale here was to provide each learning system with a representation that is easiest or most natural to process. Human participants would have likely had a harder time finding patterns in attribute-value vectors, consisting of 6 numbers, than in short video clips with moving objects. In contrast, operating on raw pixels is expected to introduce more difficulty for machine learning models in terms of disentangling representations~\cite{DBLP:conf/iclr/LazaridouHTC18}. Future work could examine whether neural nets segment visual stimuli in a similar way as humans in grounded language learning.

In conclusion, our findings shed light on the relationship between
language and language-learning systems, showing that linguistic structure is crucial for language learnability and systematic generalization in neural networks. Our results suggest that more structured languages are easier to learn, regardless of the learning system: a human, a recurrent neural network, or a large language model. Thus, generalization capabilities are heavily influenced by compositional structure, with both biological and artificial learning systems benefitting from more structured input by facilitating more systematic and transparent generalizations. In
future work, we will analyze how this learnability bias for more structure affects neural networks engaged in collaborative communication games, and test how this kind of systematic structure arises in the first place in emergent communication  simulations.
Moreover, our findings give a clear prediction that children would benefit from more structure in the linguistic input, which we will test by conducting a learnability study with children.

\section*{Methods}

\paragraph{Input Languages} The input languages with different degrees of compositional structure come from a previous communication study in which groups of
interacting participants took turns producing and guessing labels for different
dynamic scenes, creating new artificial languages over time
\cite{raviv2019larger}. Ten of the final languages created by these groups then served
as input languages for a follow-up study on language learnability with
humans~\cite{raviv2021easy2learn}. For our experiments, we used the same ten
input languages. These input languages are considered the ground truth.
Each of the ten input languages contains a set of 23 label-scene mappings. Each
scene comprises one of four different shapes moving in different directions
between 0 and 360 degrees. The languages vary in their degree of compositional structure, with structure scores ranging from $0.09$ to $0.85$.

\paragraph{Topographic Similarity to Quantify Compositional Structure} Crucially, the ten
input languages have different degrees of structure, ranging from languages
with no structure to languages with consistent, systematic grammar. Each
language has a structure score represented by topographic
similarity~\cite{DBLP:journals/alife/BrightonK06}, quantifying the degree to
which similar labels describe similar meanings. The topographic similarity is
measured as the Pearson correlation between all labels' pairwise length-normalized edit
distances and their corresponding pairwise semantic differences.
The semantic difference between two scenes is calculated as the sum of the
difference in shape and the difference in angles~\cite{raviv2021easy2learn}.
The difference in shape is zero if the two scenes contain the same shape, and
one otherwise. The difference in angles is calculated as the absolute
difference divided by 180.  The topographic similarity of a language is then
calculated as the pairwise correlation between all semantic differences and all
normalized edit distances. For a complete list of input languages and their
structure scores, see Table 3 in the SI.

\paragraph{Human Learning Data} Aside from the input languages, we use
reference data from 100 human participants learning these input
languages~\cite{raviv2021easy2learn}. The participants were different from those
who created the languages. A hundred participants, ten per input language,
engaged in repeated learning blocks consisting of passive exposure (in which
the target label-meaning mappings were presented on the screen one by one),
guessing trials (in which participants needed to pick the right scene from a
set of possible distractors), and production trials (in which participants
needed to generate a descriptive label for a target scene based on what they
had learned). During training, humans received feedback on their performance.

\paragraph{Large Language Models} For the large language models, we supplied
the full training data of the respective input language to GPT-3.5: 23 lines
consisting of shape-angle pairs in a textual format, and the corresponding
target label.  These 23 lines were followed by a single line that only
contained shape and angle but no word. GPT-3.5 was made to predict the most
likely word as completion, for which it could take into account the 23 triples
presented in the prompt.  In the memorization task, the target word appears
earlier in the prompt, which means that the perfect solution would be to simply
copy this word. In the generalization task, we gave GPT-3.5 a combination of
shape and angle not present in the training data (and not in the prompt). The
model generated the most likely descriptive word for the new shape-angle
pair.

We had to make certain technical choices when using GPT-3.5.  First, we chose a
consistent input representation (Javascript Object Notation). We do not insert
a task description to avoid potential bias. Instead we purely rely on
next-token prediction. Second, we set the sampling temperature to zero, which
controls the randomness of the generation, such that we obtain deterministic
generations. Third, we do not impose any restrictions on the characters that
can be generated but rely on its ability to detect this pattern from the
training data. Fourth, we do not feed back GPT-3.5's previous productions into
the prompt. Lastly, GPT-3.5's tokenization procedure (how text is split into
subword tokens) could have been problematic for applying it to our artificial
languages. However, we found that GPT-3.5 still reaches high memorization
performance, which suggests that tokenization is not a problem. We have
confirmed that the words of the artificial languages are tokenized as expected
with the OpenAI's
Tokenizer (\url{https://platform.openai.com/tokenizer}): falling back
to one token per character.

\paragraph{Custom Recurrent Neural Network Architecture} Our custom model
architecture (see Figure~\ref{fig:one}, right) is based on two components: a generative
component and a contrastive component.  The generative component is conditioned
on the input scene and generates a label letter by letter. The contrastive
component ensures that the matching scenes and labels are close in the
representation space and non-matching pairs are apart from each other.
For processing the sequence of letters, each component uses a
recurrent neural network, for which we use the well-known long short-term
memory (LSTM)~\cite{lstm}. In the following, we describe the input representation before
we describe the two components and their interactions.

Scenes were shown to human participants as short videos~\cite
{raviv2021easy2learn}. For the recurrent neural networks, we use a simplified
representation of the scenes.
The rationale for choosing this input representation over images is
that both humans and models receive the respective easiest possible input type
to process, allowing for a fair
comparison~\cite{firestonePerformanceVsCompetence2020,schynsDegreesAlgorithmicEquivalence2022}.
For the recurrent neural networks, we employ a one-hot encoding of the shape concatenated
with a sine and a cosine transformation of the angle.  The sine--cosine
transformation promotes a similar treatment of angles that are close to each
other, while each unique angle can be distinguished.  For example, shape~2
(between 1 and 4) moving at a 90-degree angle is converted to a vector
$(0,1,0,0,1,0)$, shape~3 with 45 degrees is converted to $(0,0,1,0,0.71,0.71)$,
and shape~4 with 135 degrees is converted to $(0,0,0,1,0.71,-0.71)$.  We refer
to the resulting 6-dimensional vector representation of the input as a scene
$\bm{x}$.
By using this input representation, we focus on the ability of systematic generalization in language
learning rather than the ability to learn disentangled representations.
If the neural networks were trained on pixel input instead, the task would be more
challenging as neural networks would need to learn disentangled
representations on the fly~\cite{DBLP:conf/iclr/LazaridouHTC18}.

Within the generative component, the input scene $\bm{x}$ is first encoded to a
latent representation $\bm{h}$ by $\textsc{Encoder}$, a feedforward network (we use
a multilayer perceptron with one hidden layer), such that we obtain a latent
representation $\bm{h} = \textsc{Encoder}(\bm{x})$.  This latent representation $\bm{h}$
is then used as the initial state of the recurrent neural network
$\textsc{Writer}$. The $\textsc{Writer}$ sequentially produces a sequence of letters,
\ie, a label, as output. This $\textsc{Writer}$ consists of three modules: an
input embedding for previously produced characters, an LSTM cell, and an output
layer that produces the next letter.

For the contrastive component, we use another recurrent module $\textsc{Reader}$
that reads a label $\bm{m}$ sequentially (\ie, letter by letter) while updating
its state. As for the $\textsc{Writer}$, we again use an LSTM.  A fully-connected
layer transforms the final state into a latent representation $\bm{z}$, such that
$\bm{z} = \textsc{Reader}(\bm{m})$, where $m$ is the input label. The reading component
is used for contrastive learning, \ie, they are trained so that the hidden
representation of the label $\bm{z}$ matches the representation of the
corresponding scene $\bm{h} = \textsc{Encoder}(\bm{x})$, which is used as the initial
hidden state of the generative $\textsc{Writer}$ module.

To ensure that the contrastive training procedure affects the generative
component, we couple the two components: First, the
embedding (\ie, the mapping between the agent's alphabet and the first latent
representation) parameters are shared between the input layer of $\textsc{Reader}$,
the input layer of $\textsc{Writer}$, and the output layer of $\textsc{Writer}$. Second,
the same encoder module is used in both the generative and the contrastive
components (see Figure~\ref{fig:one}).

The output dimension of $\textsc{Encoder}$, the hidden state sizes of $\textsc{Reader}$
and $\textsc{Writer}$, and the embedding size are all set to 50. A sensitivity
analysis of the hidden size on the dependent variables of interest is provided
in Figs.~18--20 in the SI\@. Similarly to \citet{nakkiran2021deep}, larger
hidden sizes led to a faster increase in memorization and generalization.

\paragraph{Training Procedure} We train the recurrent neural networks for
multiple training rounds as in the experiments with human
participants~\cite{raviv2021easy2learn}. Each training round consists of three
blocks: exposure, guessing, and production block, described in detail in the
following.  As typical in neural network training, we train the network with
backpropagation and stochastic gradient descent, where the gradient is
estimated based on a small number of examples
(minibatches)~\cite{rumelhart_learning_1986,goodfellow_deep_2016}.
The batch size, which also determines the number of distractors, is set to 5,
reflecting human short-term memory constraints~\cite{cowan2001magical}.  Only in
the guessing block, we set the batch size of 1 and use the same distractors as
in the experiments with human participants, instead of other exemplars from the
same batch.

In the exposure block, human participants were exposed to scenes with the
corresponding target labels.  Therefore, we train the deep learning models
using a loss function with two terms: a generative and a contrastive loss term.
The generative loss, $\mathcal{L}_\mathrm{gen}$, is a token-wise cross-entropy with the
ground-truth label of the original language. The contrastive loss,
$\mathcal{L}_\mathrm{con}$, promotes similar latent representations of scenes and
labels that correspond to each other and contrasts representations that do not.
Specifically, we use the normalized temperature-scaled cross-entropy loss
(NTXent)~\cite{simclr}.  We use other scenes in the same batch as distractors
for the contrastive loss term.  The final loss function is $\mathcal{L} =
\mathcal{L}_\mathrm{gen} + \alpha_\mathrm{con} \mathcal{L}_\mathrm{con}$. The factor
$\alpha_\mathrm{con}$ determines the relative weight of the loss terms. For the
main experiment, we use $\alpha_\mathrm{con}=0.1$. A sensitivity analysis using
other values for $\alpha_\mathrm{con}$ is provided in Figs.~21-23 of the SI.

In the guessing block, we use the same loss function as in the exposure block.
The contrastive loss term $\mathcal{L}_\mathrm{con}$ mirrors the task in which human
participants had to select the correct scene against the distractors given a
label.  The generative loss term $\mathcal{L}_\mathrm{gen}$ is used so that the model
does not ``forget'' how to generate~\cite{catforget}. Notably, the guessing task itself could be also carried out by having the models generate a descriptive label for each scene and then select the closest one to the given label in terms of edit distance. However, we opted for optimizing shared parameters through a contrastive loss to ensure that the guessing task would also have an effect on the production task (and vice-versa).

In more detail, the latent representation $\bm{z} = \textsc{Encoder}(\bm{x})$ of the
scene $\bm{x}$ should be closest to the latent representation $\bm{z}^\prime =
\textsc{Reader}(\bm{m})$ of the corresponding label $\bm{m}$.  The difference from
exposure training is that in the guessing block, we use the identical
distractors used in experiments with humans, whereas, in the exposure block, we
use the other scenes from the same batch. The trajectory of guessing accuracy
during training is shown in Fig. 12 in the SI.

In the production block, a scene was presented to human participants, who had
to produce a label.  We again use the same generative loss as in the previous
block, $\mathcal{L}_\mathrm{gen}$, to model the production block.  In the production
block, however, we omit the contrastive loss term and train only on generation.
Thus, the loss function for the production block is $\mathcal{L} = \mathcal{L}_\mathrm{gen}$.

The parameters are randomly initialized by He
initialization~\cite{he2015delving}, the default initialization method in
PyTorch~\cite{paszke_pytorch_2019}.
We employ the widely used Adam optimizer~\cite{kingma2015adam} to carry
out the optimization of the  loss function with the default learning rate of
$10^{-3}$.  As common in machine learning, we have to make certain decisions
about the neural network architecture design, optimization procedure, and
hyperparameters. All these decisions may impact the results. However, we have
varied relevant hyperparameter settings and found that the results are
robust and do not dependent on specific settings of the hyperparameters (see Figs.~24 and 25 in the SI).

\paragraph{Measures}\label{sub:measures}
Production similarity measures the overlap between two sets of labels. It is computed as one minus the normalized edit distance between pairs of labels.
For our analysis, we use production similarity once to quantify
the similarity between the generated labels and the ground truth of the input languages, and once to quantify the similarity of labels generated by neural network agents with labels produced by human learners.
For example, a recurrent neural network that produced 'muif-a' for shape 3 moving in direction 360 degrees would have a high production similarity to the majority of human participants who produced 'muif-i'.

The generalization score measures the degree of systematicity during the
generalization test~\cite{raviv2021easy2learn}.  We take two sets of scenes: a
training set, on which the agents were trained, and a test set, on which the
agents were not trained.  We then do the following for each agent.  First, we
take two sets of labels: one previously generated for each training scene by
the agent and another that we let the agent generate for each test scene.
Second, the difference between train and test scenes is measured by pairwise
semantic difference. Semantic difference is calculated as topographic similarity. Third, the difference between train and test labels is
measured by pairwise normalized edit distance.  Finally, we compute the Pearson
correlation between these two differences across all scenes.  Then, we take the
average correlation coefficient across all agents as the generalization score.

The convergence score measures the similarity in the generalization test
between agents that learned the same language.  We take the test set on
which the agents have not been trained and let each agent produce a label for
each scene.  We compute the pairwise normalized edit distance between all
generated labels per scene so that if we have $n$ test scenes and $k$ agents,
we compute $n \cdot \frac{k(k-1)}{2}$ distances.  We then compute the average
distance across both scenes and labels and take one minus the average distance
as the convergence score.  Therefore, if all agents produce the same label for
each test scene, we would get a convergence score of 1. Conversely, if each agent produced a different label for the same scene, the convergence score would be zero

\paragraph{Statistical Analyses} We trained 100 differently-initialized neural
network models over 100 rounds for each of the ten input languages. The testing
in each round consisted of 23 memorization and 13 generalization examples.
This makes a total of 2.3M memorization and 1.3M generalization test results
subject to statistical analyses.  Significance was tested using linear
mixed-effects models, as implemented in the Python package
statsmodels~\cite{seabold2010statsmodels}, for production similarity
(LME~1\xspace), generalization score (LME~3\xspace), generalization convergence
(LME~4\xspace), as well as production similarity to humans in
memorization (LME~2\xspace) and generalization (LME~5\xspace). We use
the structure score and the logarithmized round number in all measures as a
fixed effect.  The number of rounds was logarithmized following scaling laws of
neural language models~\cite{kaplanScalingLawsNeural2020}. Both the structure
score and the logarithmized round number were centered and scaled We consider
two random effects: the random seed for initialization (which also determines
the input language) and the specific scene.  For LME~5\xspace,
scaling the log-transformed round number to unit variance hindered convergence,
so the log rounds were only centered. The full results of the statistical
models are provided in Tab.~4 in the SI, with partial regression plots shown
in Figs.~13--17. In Tab.~5 in the SI, we provide an  additional analysis of
production similarity to ground truth at rounds 10, 40, 70, and 100.


\section*{Code Availability}
The code for reproducing our experiments is available on GitHub \url{https://github.com/lgalke/easy2deeplearn}.

\bibliography{main,extra}
\bibliographystyle{unsrtnat}

\section*{Acknowledgments} We thank
Dota Tianai Dong, Koen de Reus,
Yosef Prat, Tal Simon, Willem Zuidema, Tessa Verhoef, Mitja Nikolaus, Marieke Woensdregt, and
Adam Kohan for their comments and discussions. We thank Shinje Wu and Marianne
de Heer Kloots for sharing their data.



\appendix

\newpage
\section{Supplementary Information}

\newcommand\modelMemProdSim[0]{LME~1\xspace}
\newcommand\modelMemProdSimHumans[0]{LME~2\xspace}

\newcommand\modelRegGenScore[0]{LME~3\xspace}
\newcommand\modelRegConvScore[0]{LME~4\xspace}
\newcommand\modelRegProdSimHumans[0]{LME~5\xspace}

\newcommand\modelPointwiseMemProdSim{LME~6\xspace} 

\newcommand\scalefig{1.0}

\section*{Details of the input languages}
 Table~\ref{tab:structurescores}.

\begin{table}[p!]
    \centering
    \begin{tabular}{cccc}
        \toprule
        \textbf{Input Language} & \textbf{Structure Score} & \textbf{Ambiguity \%} & \textbf{Structure Bin}\\
        \midrule
         S1& 0.09 & 0  & 1\\
         B1& 0.07 & 0  & 1\\
         S2& 0.25 & 0.35  & 2\\
         B2& 0.35 & 0.09  & 2\\
         S3& 0.59 & 0.13  & 3\\
         B3& 0.58 & 0.17  & 3\\
         S4& 0.79 & 0  & 4\\
         B4& 0.69 & 0  & 4\\
         S5& 0.84 & 0  & 5\\
         B5& 0.85 & 0  & 5\\
         \bottomrule
    \end{tabular}
    \caption{Structure scores of the input languages}
    \label{tab:structurescores}
\end{table}

\section*{Extended Results}
\subsection*{Production Similarity to Ground Truth during Memorization}
\begin{figure}[hp]
    \centering
    \includegraphics[width=\scalefig\textwidth]{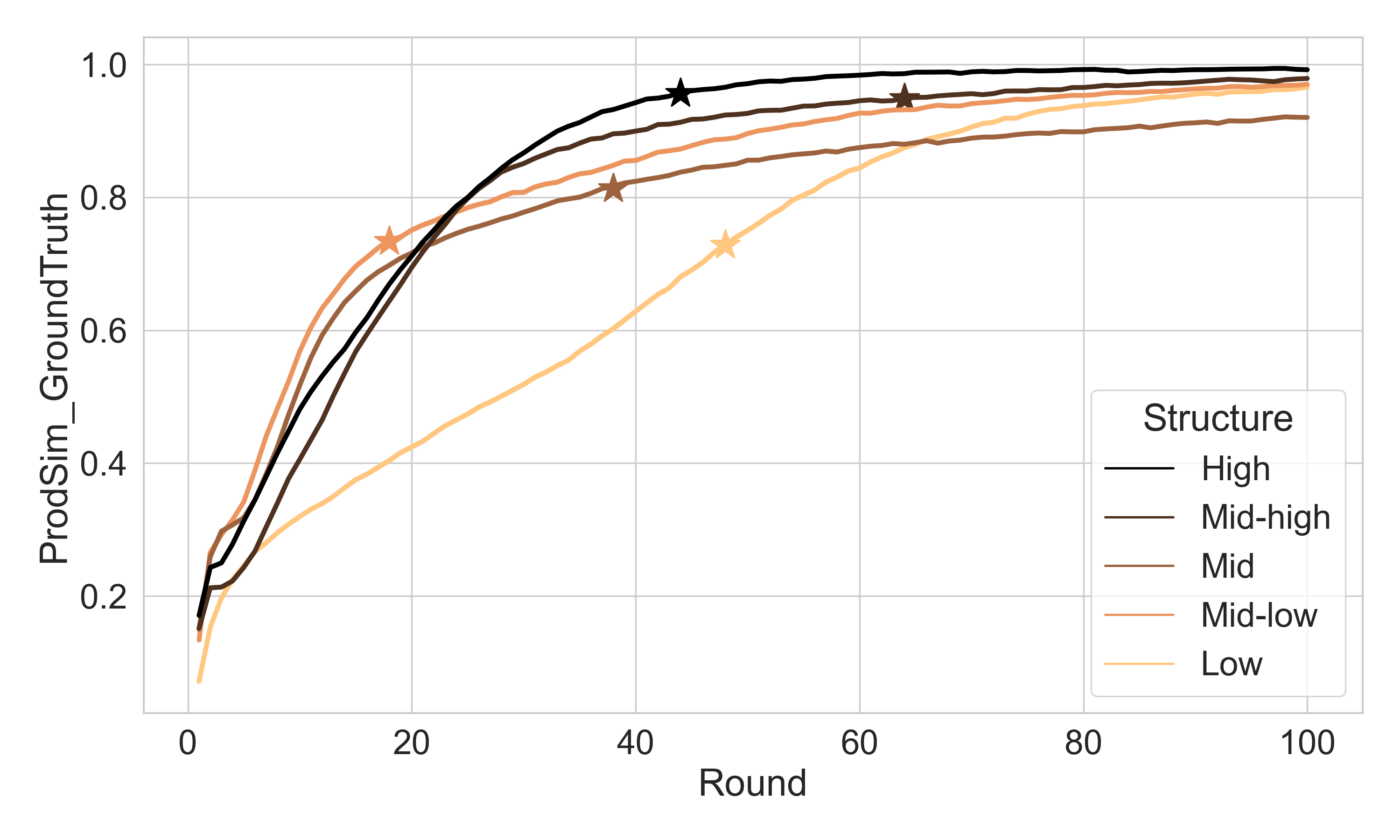}
    \caption{Production Similarity to the ground truth of the input language as a function of round number. Color indicates the degree of structure (darker means higher). Stars indicate where neural network agents exceed human performance.}
    \label{fig:reg-prodsim:relplot}
\end{figure}

Figure~\ref{fig:reg-prodsim:relplot}

\subsection*{Production Similarity to Humans during
Memorization}
\begin{figure}[hp]
    \centering
    \includegraphics[width=\scalefig\textwidth]{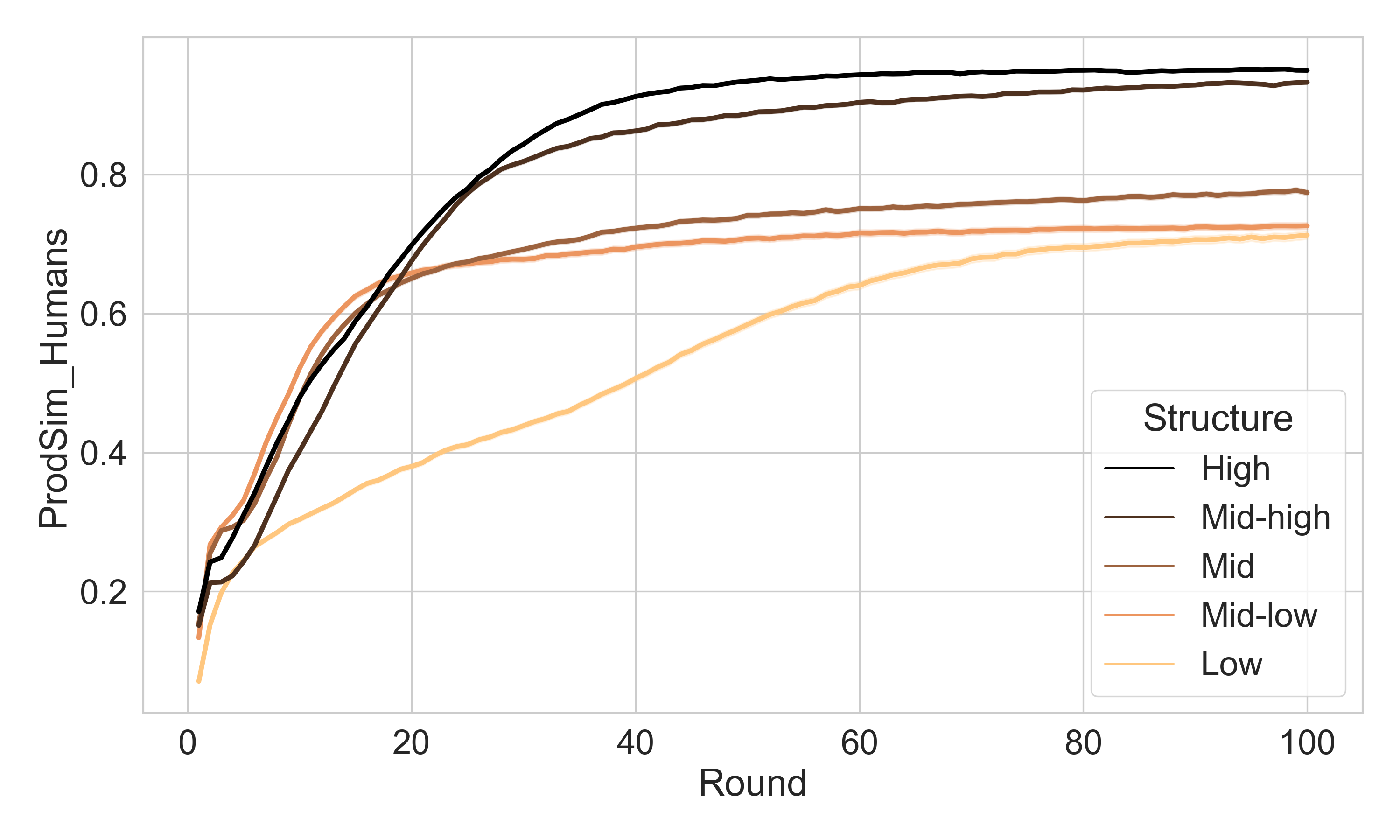}
    \caption{Production similarity to humans learning the same input language during memorization. Color indicates the degree of structure (darker means higher).}
    \label{fig:mem-prodsim-humans:relplot}
\end{figure}

Figure~\ref{fig:mem-prodsim-humans:relplot}

\subsection*{Systematicity during Generalization}
\begin{figure}[hp]
    \centering
    \includegraphics[width=\scalefig\textwidth]{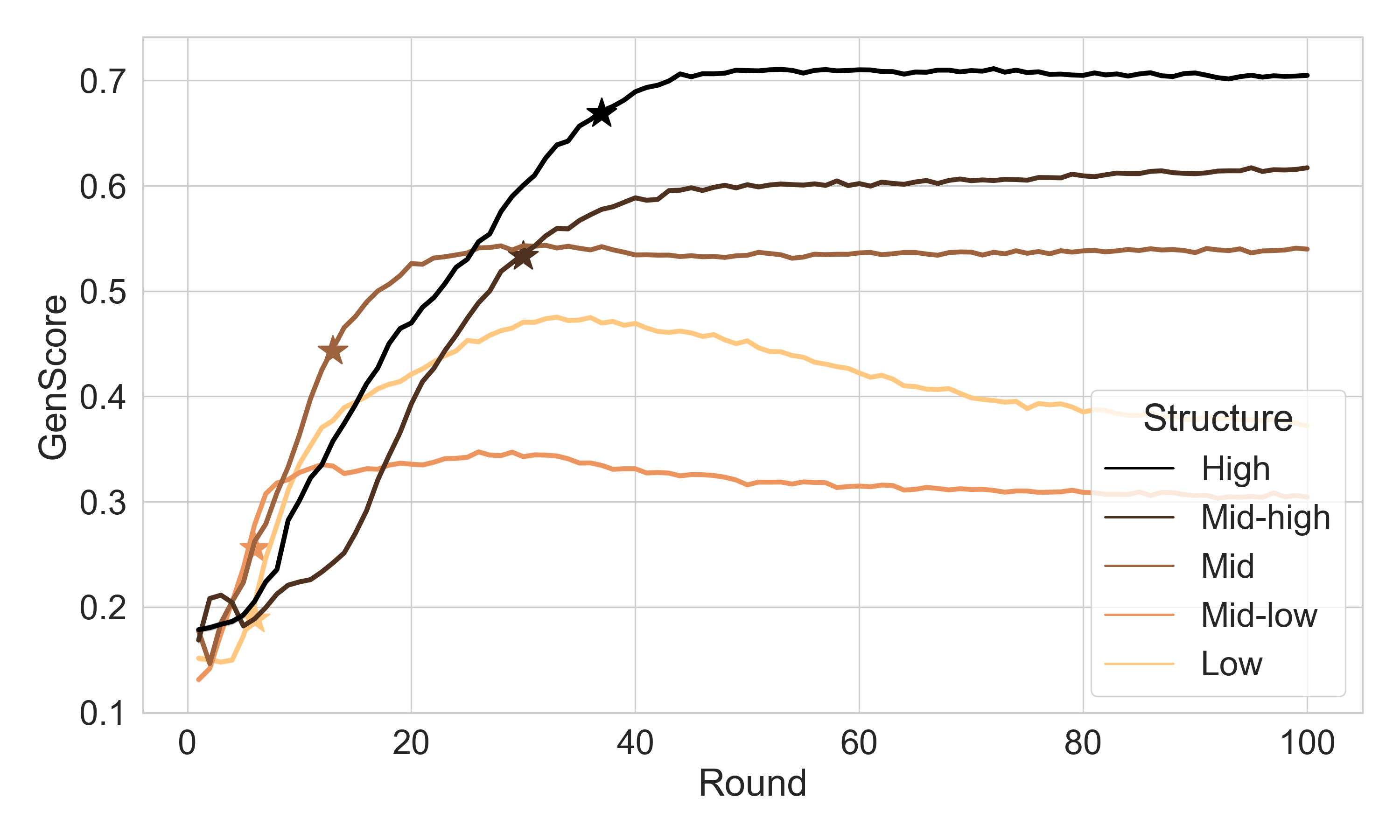}
    \caption{Generalization score as a function of round number. Color represents the degree of structure. Stars indicate where neural network agents exceed human performance.}
    \label{fig:reg-genscore:relplot:prenorm}
\end{figure}

Figure~\ref{fig:reg-genscore:relplot:prenorm}

\subsection*{Convergence Score}
\begin{figure}
    \centering
    \includegraphics[width=\scalefig\textwidth]{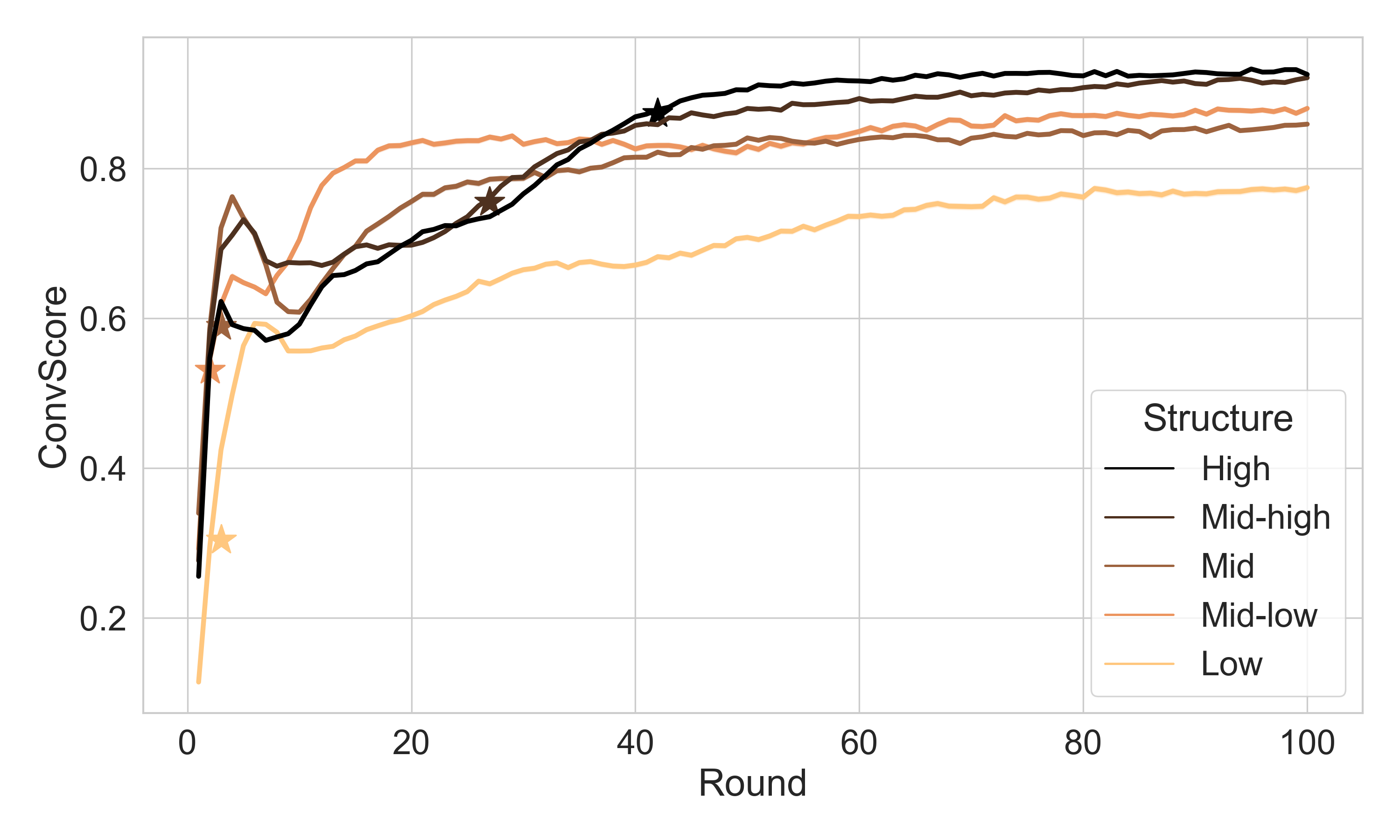}
    \caption{Convergence score as a function of round number. Color indicates the degree of structure (darker means higher). Stars indicate where neural network agents exceed human performance.}
    \label{fig:reg-convscore:relplot}
\end{figure}

 Figure~\ref{fig:reg-convscore:relplot}

\subsection*{Production Similarity to Humans during Generalization}
\begin{figure}
    \centering
    \includegraphics[width=\scalefig\textwidth]{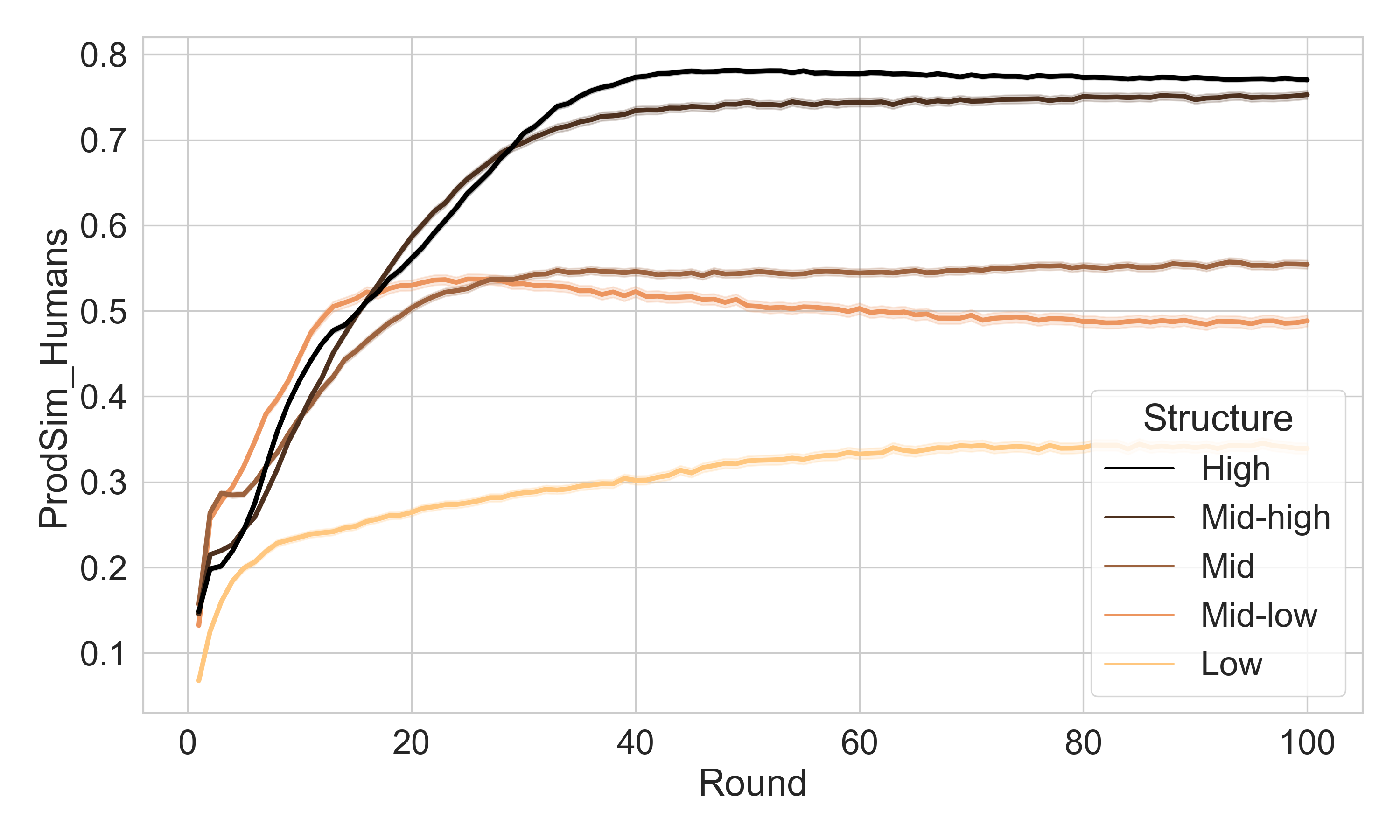}
    \caption{Production Similarity to humans during generalization. Color indicates the degree of structure (darker means higher).}
    \label{fig:reg-prodsim-humans:relplot}
\end{figure}

Figure~\ref{fig:reg-prodsim-humans:relplot}

\subsection*{Binary Accuracy with respect to Ground Truth during Memorization}
\begin{figure}
    \centering
    \includegraphics[width=\scalefig\textwidth]{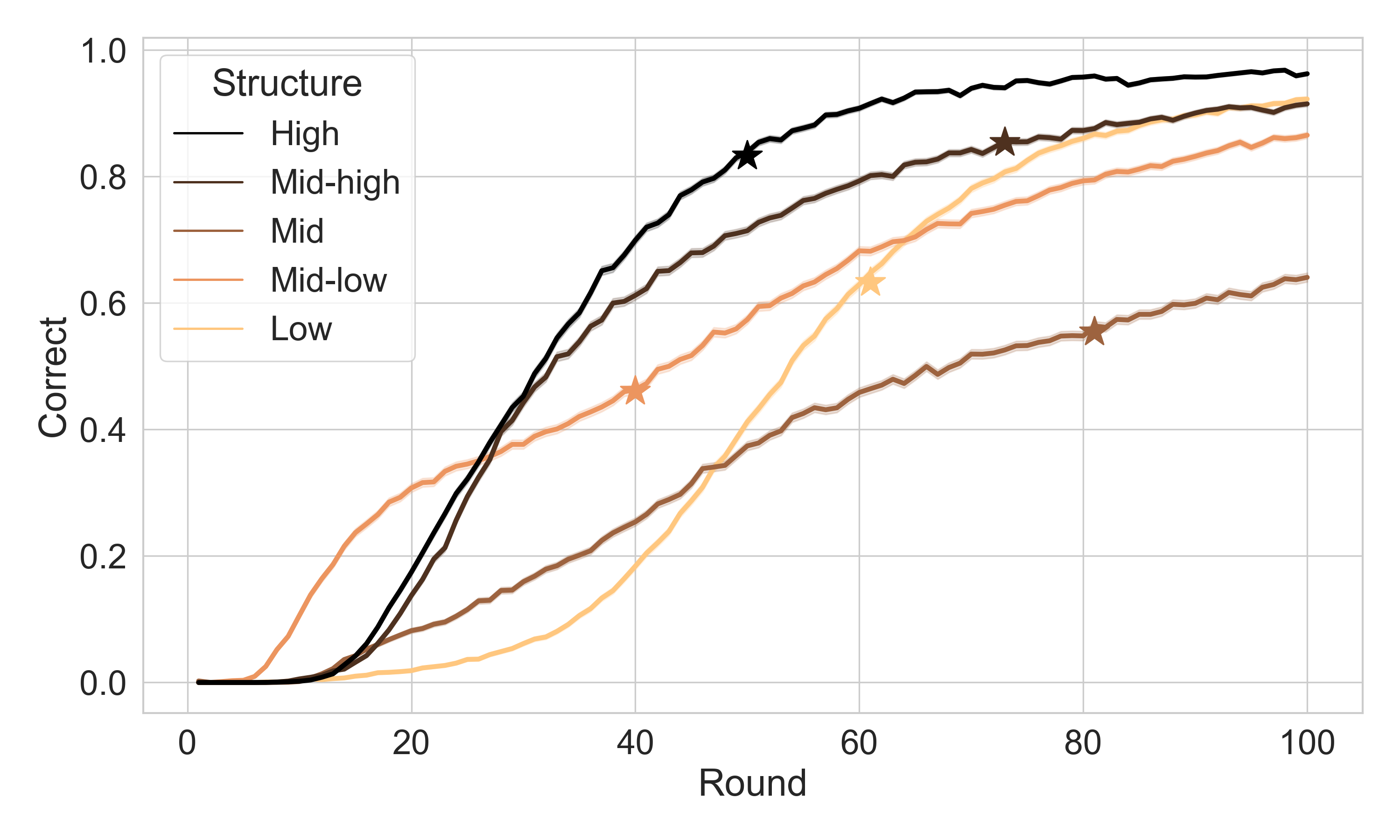}
    \caption{Binary Accuracy with respect to ground truth of input languages as a function of round number. To compute binary accuracy, we compare the labels produced by the neural agents with the ground truth label of the input language and each label receives a score of one if it is exactly the same as the ground truth and zero otherwise. These boolean scores are then averaged to obtain binary accuracy.
    Color indicates the degree of structure (darker means higher). Stars indicate where neural network agents exceed human performance.}
    \label{fig:mem-acc:relplot}
\end{figure}

 Figure~\ref{fig:mem-acc:relplot}

\subsection*{Guessing Accuracy}
\begin{figure}
    \centering
    \includegraphics[width=\scalefig\textwidth]{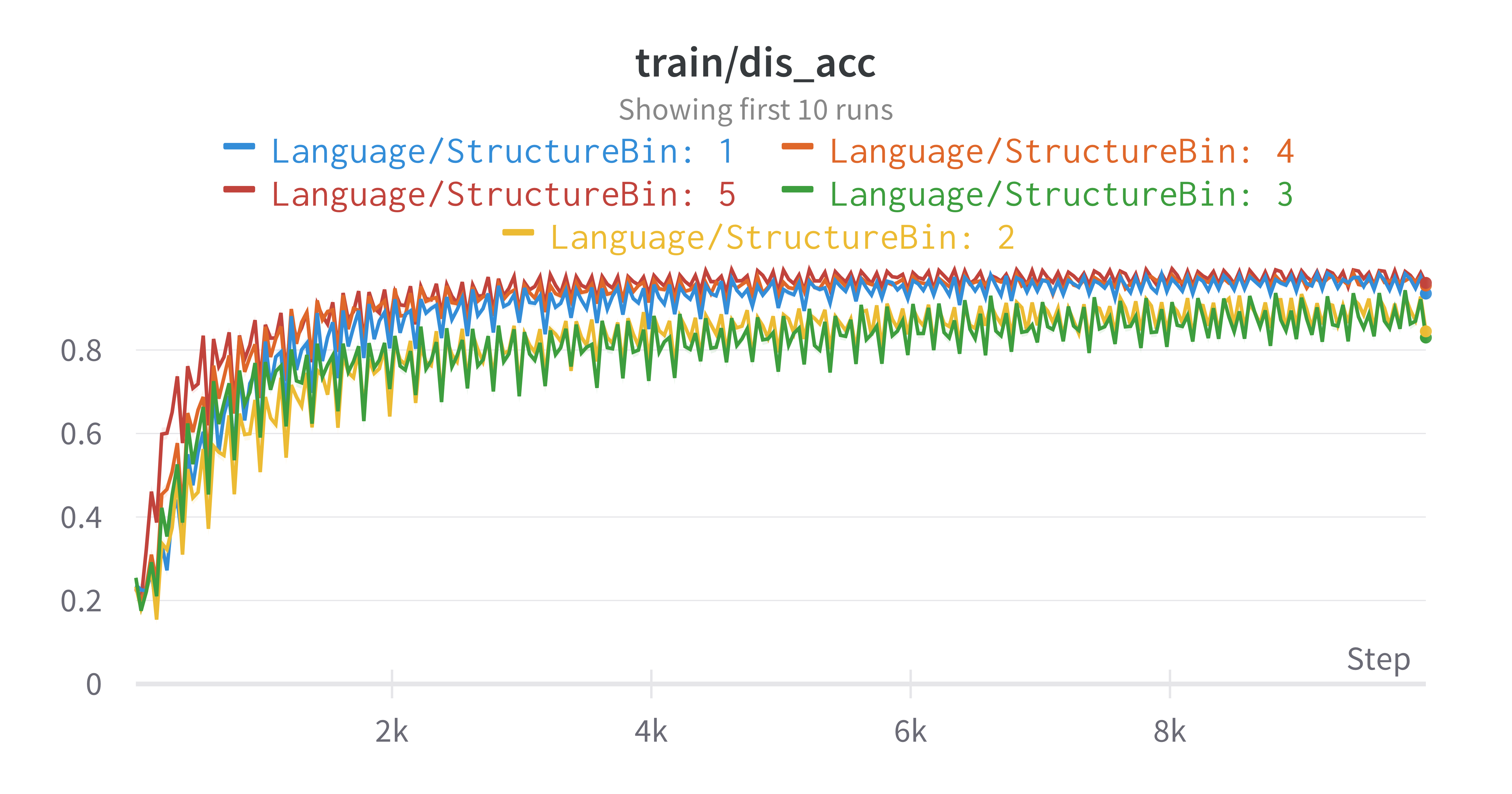}
    \caption{Guessing Accuracy: selecting the right scene among distractors within the contrastive training objective during training. Guessing accuracy was calculated during training with active dropout since it was part of the training phase.} \label{fig:guessing-accuracy}
\end{figure}

Figure~\ref{fig:guessing-accuracy}

\section*{Statistical Analyses of the Results }\label{sec:stats:multiple}

Table~\ref{tab:stats} shows the results of the statistical models \modelMemProdSim, \modelMemProdSimHumans, \modelRegGenScore, \modelRegConvScore, \modelRegProdSimHumans, and \modelPointwiseMemProdSim.

\subsection*{\modelMemProdSim: Production Similarity during Memorization}
\begin{figure}
    \centering
    \includegraphics[width=\scalefig\textwidth]{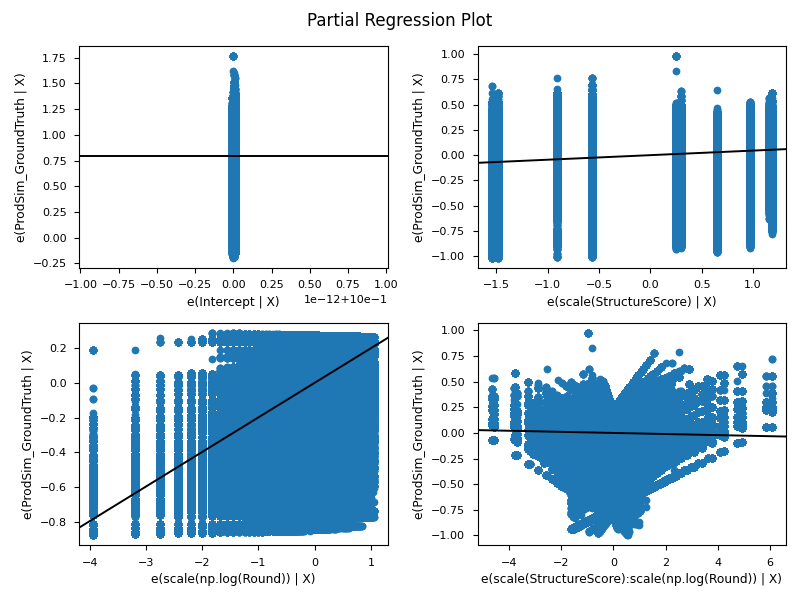}
    \caption{Partial regression plots of \modelMemProdSim : Production Similarity to ground truth during memorization}
    \label{fig:mem-prodsim:partregress}
\end{figure}

The dependent variable is the production similarity to the ground truth in the memorization test rounds rounds.
Figure~\ref{fig:mem-prodsim:partregress} shows the partial regression plots.

\subsection*{\modelMemProdSimHumans: Production Similarity to Humans during Memorization}
\begin{figure}[p]
    \centering
    \includegraphics[width=\scalefig\textwidth]{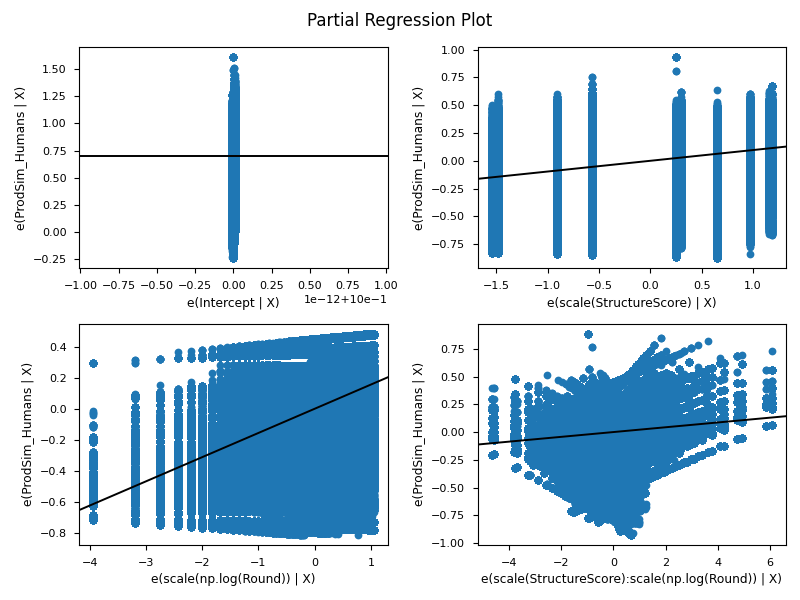}
    \caption{Partial regression plots of \modelMemProdSim: Production Similarity to humans during Memorization}
    \label{fig:mem-prodsim-humans:partregress}
\end{figure}

The dependent variable is the production similarity to human participants in the memorization test rounds.
Figure~\ref{fig:mem-prodsim-humans:partregress} shows the partial regression plots.

\subsection*{\modelRegGenScore: Systematicity during Generalization}
\begin{figure}[p]
    \centering
    \includegraphics[width=\scalefig\textwidth]{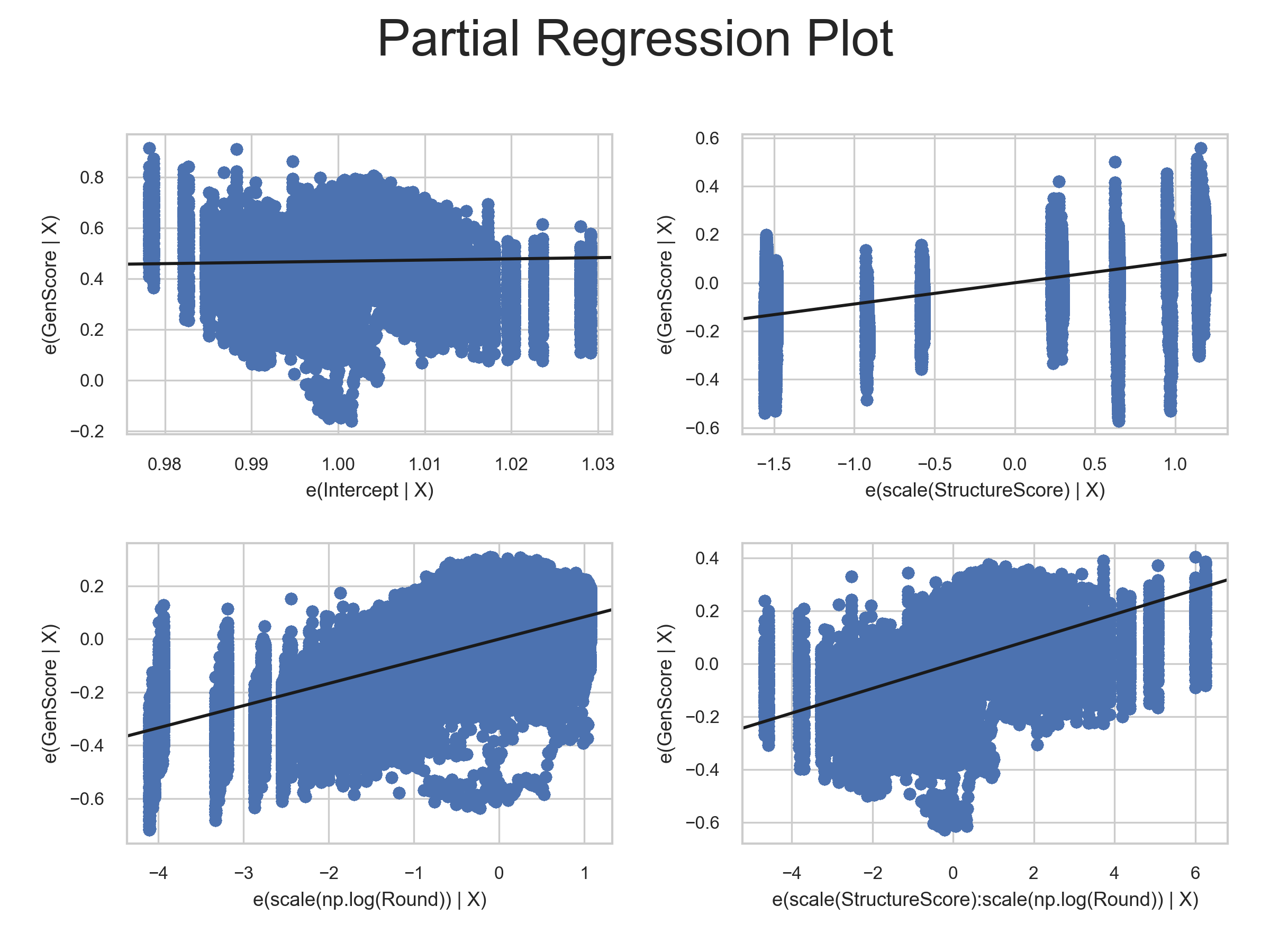}
    \caption{Partial regression plots of \modelRegGenScore : Generalization Systematicity}
    \label{fig:reg-genscore:partregess}
\end{figure}

The dependent variable is the generalization score in the generalization test rounds.
Figure~\ref{fig:reg-genscore:partregess} shows the partial regression plots.

\subsection*{\modelRegConvScore: Convergence Score during Generalization}
\begin{figure}[p]
    \centering
    \includegraphics[width=\scalefig\textwidth]{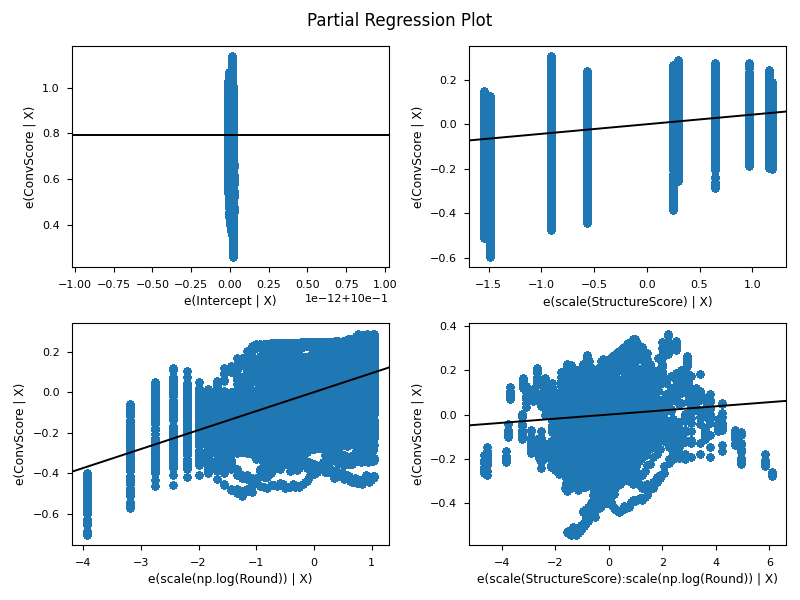}
    \caption{Partial regression plots of \modelRegConvScore: Convergence Score (during Generlization)}
    \label{fig:reg-convscore:partregress}
\end{figure}

The dependent variable is the convergence score in the generalization test rounds.
Figure~\ref{fig:reg-convscore:partregress} shows the partial regression plots.

\subsection*{\modelRegProdSimHumans: Production Similarity to Humans during Generalization}
\begin{figure}[p]
    \centering
    \includegraphics[width=\scalefig\textwidth]{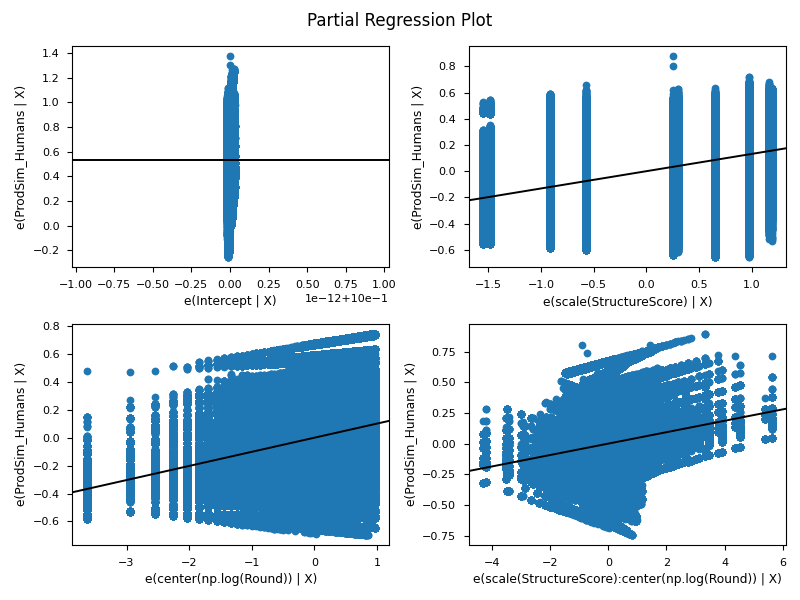}
    \caption{Partial regression plots of \modelRegProdSimHumans: Production similarity to humans during generalization}
    \label{fig:reg-prodsim-humans:partregress}
\end{figure}

The dependent variable is the production similarity to human participants in the generalization test rounds.
Figure~\ref{fig:reg-prodsim-humans:partregress} shows the partial regression plots.

\subsection*{\modelPointwiseMemProdSim{}: Production Similarity to Ground Truth at Specific Rounds}
Table~\ref{tab:pointwise-mem-prodsim} shows the results of the statistical models for the production similarity to ground truth during memorization at specific rounds: 10, 40, 70, and 100.

\begin{table}
\footnotesize
\caption{Linear Mixed-Effects Regression Results. \modelMemProdSim: Production similarity to ground truth during memorization. \modelMemProdSimHumans: Production similarity to Humans during Generalization. \modelRegGenScore: Generalization Score. All tests are two-sided.}\label{tab:stats}
\begin{center}
\begin{tabular}{lrrrrrr}
\hline
               \textbf{\modelMemProdSim}: Production Similarity to Ground Truth                            &  Coef. & Std.Err. &        z & P$> |$z$|$ & [0.025 & 0.975]  \\
\hline
Intercept                                  &  0.797 &    0.001 & 1110.436 &       0.000 &  0.796 &  0.798  \\
scale(StructureScore)                      &  0.045 &    0.001 &   62.865 &       0.000 &  0.044 &  0.047  \\
scale(np.log(Round))                       &  0.199 &    0.000 & 2060.282 &       0.000 &  0.199 &  0.199  \\
scale(StructureScore):scale(np.log(Round)) & -0.005 &    0.000 &  -54.978 &       0.000 & -0.005 & -0.005  \\
\hline
\hline
               \textbf\modelMemProdSimHumans: Prod. Sim.\ to Humans during Memorization                           & Coef. & Std.Err. &        z & P$> |$z$|$ & [0.025 & 0.975]  \\
\hline
Intercept                                  & 0.701 &    0.001 &  517.888 &       0.000 &  0.698 &  0.704  \\
scale(StructureScore)                      & 0.097 &    0.001 &   71.429 &       0.000 &  0.094 &  0.099  \\
scale(np.log(Round))                       & 0.156 &    0.000 & 1504.189 &       0.000 &  0.155 &  0.156  \\
scale(StructureScore):scale(np.log(Round)) & 0.022 &    0.000 &  208.708 &       0.000 &  0.021 &  0.022  \\
\hline
\hline
         \textbf{\modelRegGenScore}: Generalization Score                                  & Coef. & Std.Err. &        z & P$> |$z$|$ & [0.025 & 0.975]  \\
\hline
Intercept                                  & 0.468 &    0.001 &  790.838 &       0.000 &  0.467 &  0.469  \\
scale(StructureScore)                      & 0.088 &    0.001 &  148.901 &       0.000 &  0.087 &  0.089  \\
scale(np.log(Round))                       & 0.084 &    0.000 & 1281.568 &       0.000 &  0.084 &  0.084  \\
scale(StructureScore):scale(np.log(Round)) & 0.046 &    0.000 &  703.483 &       0.000 &  0.046 &  0.046  \\
\hline
\hline
      \textbf{\modelRegConvScore}: Convergence Score                                     & Coef. & Std.Err. &        z & P$> |$z$|$ & [0.025 & 0.975]  \\
\hline
Intercept                                  & 0.792 &    0.001 &  900.121 &       0.000 &  0.790 &  0.794  \\
scale(StructureScore)                      & 0.043 &    0.001 &   49.027 &       0.000 &  0.041 &  0.045  \\
scale(np.log(Round))                       & 0.094 &    0.000 & 1220.090 &       0.000 &  0.094 &  0.094  \\
scale(StructureScore):scale(np.log(Round)) & 0.009 &    0.000 &  121.740 &       0.000 &  0.009 &  0.010  \\
\hline
\hline
      \textbf{\modelRegProdSimHumans}: Prod. Sim.\ to Humans during Generalization                                      & Coef. & Std.Err. &       z & P$> |$z$|$ & [0.025 & 0.975]  \\
\hline
Intercept                                   & 0.529 &    0.002 & 280.903 &       0.000 &  0.525 &  0.533  \\
scale(StructureScore)                       & 0.132 &    0.002 &  70.280 &       0.000 &  0.129 &  0.136  \\
center(np.log(Round))                       & 0.101 &    0.000 & 749.746 &       0.000 &  0.101 &  0.101  \\
scale(StructureScore):center(np.log(Round)) & 0.046 &    0.000 & 344.287 &       0.000 &  0.046 &  0.047  \\
\hline
\end{tabular}
\end{center}
\end{table}

\begin{table}
\caption{Linear Mixed-Effects Regression Results: Production Similarity to Humans during Memorization at a Fixed Round Number. All tests are two-sided.}\label{tab:pointwise-mem-prodsim}
\begin{center}
\begin{tabular}{lrrrrrr}
\hline
    \textbf{Round 10}                  & Coef. & Std.Err. &        z & P$> |$z$|$ & [0.025 & 0.975]  \\
\hline
Intercept             & 0.459 &    0.000 & 1009.015 &       0.000 &  0.458 &  0.459  \\
scale(StructureScore) & 0.024 &    0.001 &   16.470 &       0.000 &  0.021 &  0.027  \\
\hline
\hline
\textbf{Round 40}                      & Coef. & Std.Err. &        z & P$> |$z$|$ & [0.025 & 0.975]  \\
\hline
Intercept             & 0.830 &    0.000 & 2370.225 &       0.000 &  0.829 &  0.831  \\
scale(StructureScore) & 0.094 &    0.001 &   78.439 &       0.000 &  0.091 &  0.096  \\
\hline
\hline
     \textbf{Round 70}                 & Coef. & Std.Err. &         z & P$> |$z$|$ & [0.025 & 0.975]  \\
\hline
Intercept             & 0.936 &    0.000 & 19666.003 &       0.000 &  0.936 &  0.936  \\
scale(StructureScore) & 0.021 &    0.001 &    23.322 &       0.000 &  0.020 &  0.023  \\
\hline
\hline
      \textbf{Round 100}                & Coef. & Std.Err. &         z & P$> |$z$|$ & [0.025 & 0.975]  \\
\hline
Intercept             & 0.965 &    0.000 & 31871.611 &       0.000 &  0.965 &  0.965  \\
scale(StructureScore) & 0.005 &    0.001 &     7.725 &       0.000 &  0.004 &  0.007  \\
\hline

\end{tabular}
\end{center}
\end{table}

\section*{Sensitivity to Hyperparameters}\label{app:sensitivity-to-hparams}
We found the training robust to most experimental configurations, such as learning rate, number of layers, and whether the parameters are shared between reader and writer models.
However, one particular hyperparameter that affects the model capacity has a substantial effect on the learning speed: the size of the hidden layers, for which we provide a sensitivity analysis in the following.
Subsequently, we also provide a sensitivity analysis of the scaling factor $\alpha$ for the contrastive loss term $\mathcal{L}_\mathrm{con}$.

\subsection*{Sensitivity to Hidden Layer Size}
\begin{figure}
    \centering
    \includegraphics[width=\scalefig\textwidth]{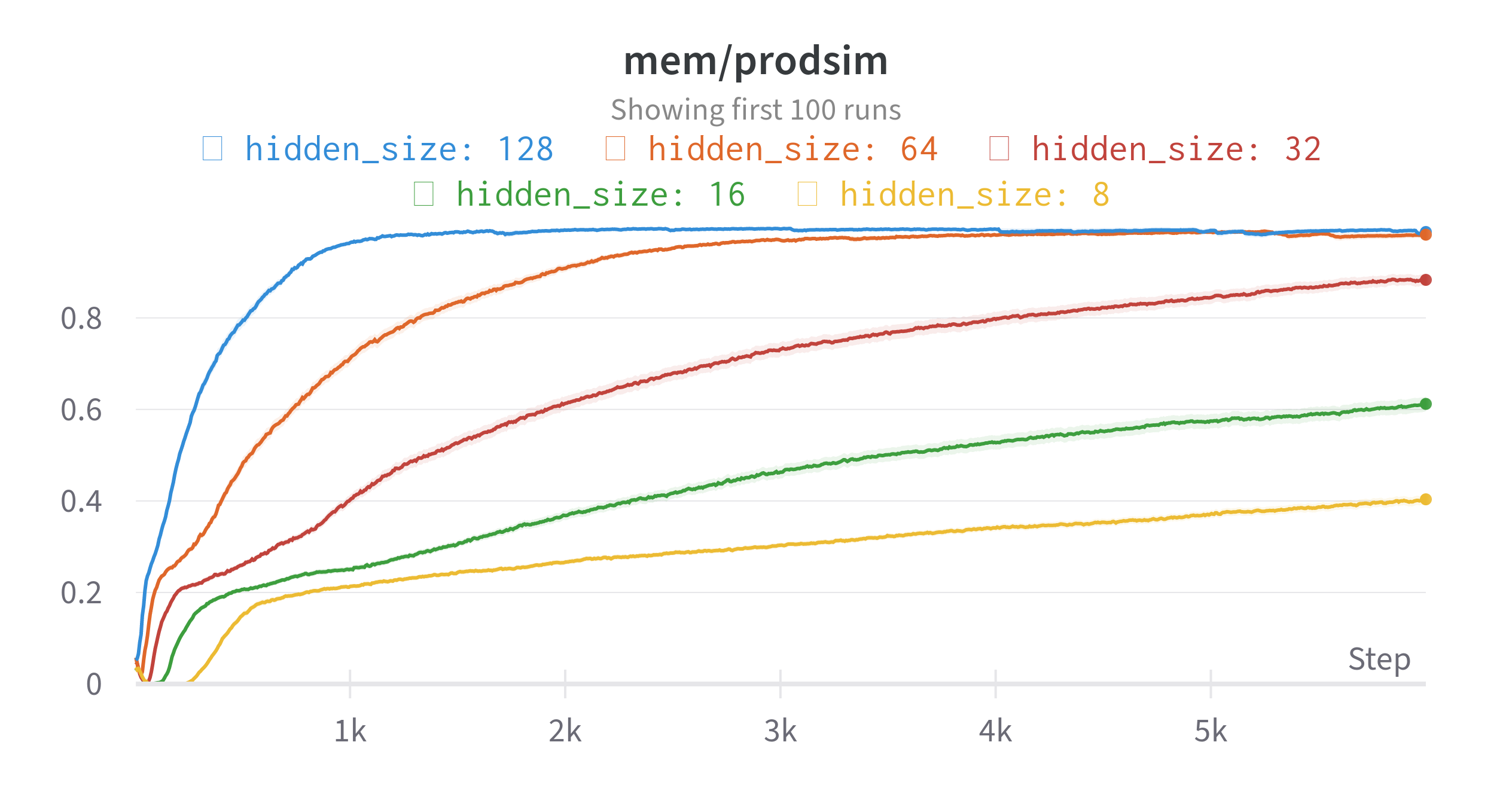}
    \caption{Average production similarity to ground truth during memorization across input languages as a function of the size of the neural networks' hidden layers.}
    \label{fig:hiddensize:mem-prodsim}
\end{figure}

\begin{figure}[p]
    \centering
    \includegraphics[width=\scalefig\textwidth]{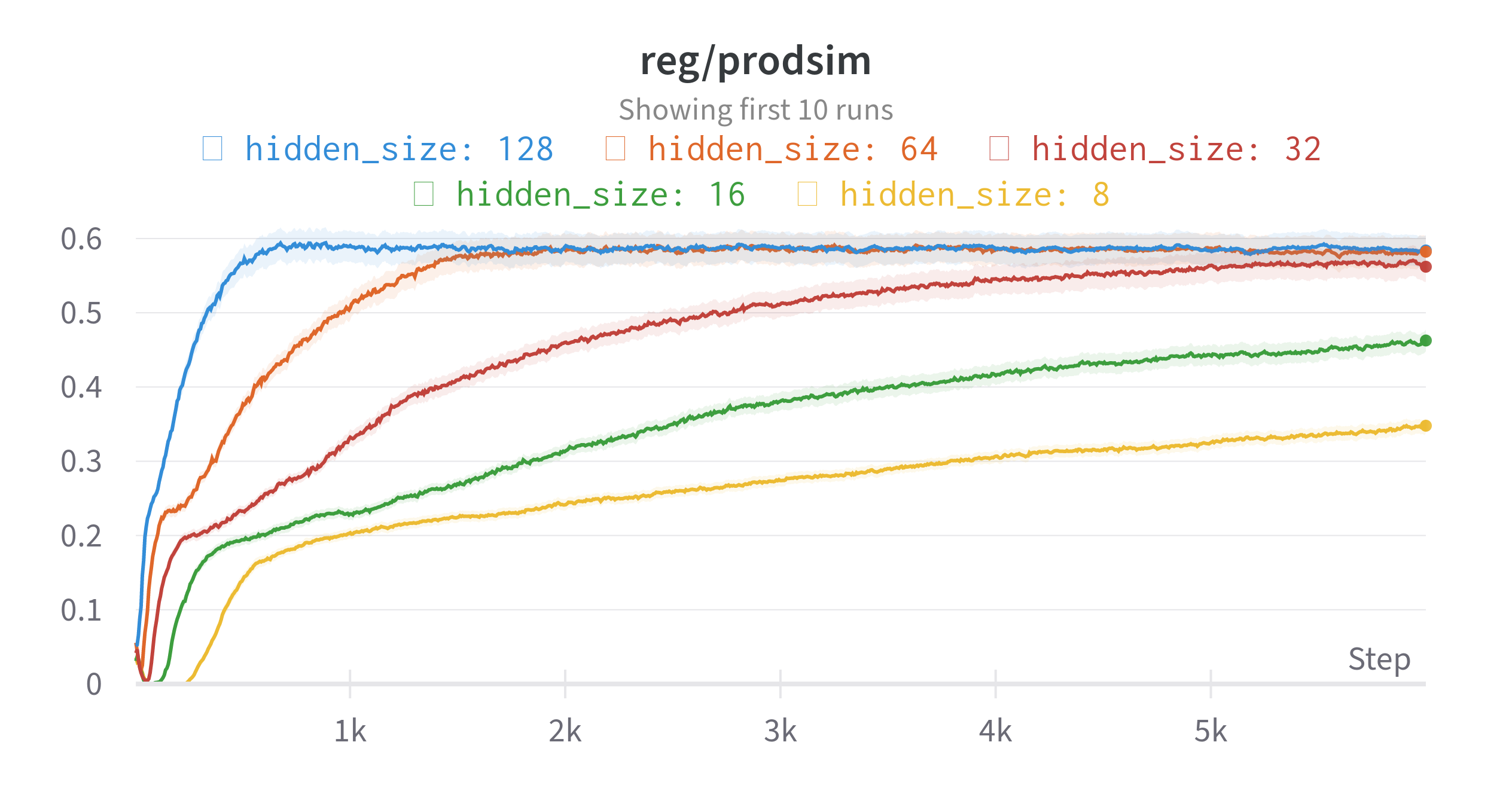}
    \caption{Average production similarity to humans across input languages as a function of the size of the neural networks' hidden layers.}
    \label{fig:hiddensize:reg-prodsim}
\end{figure}

\begin{figure}[p]
    \centering
    \includegraphics[width=\scalefig\textwidth]{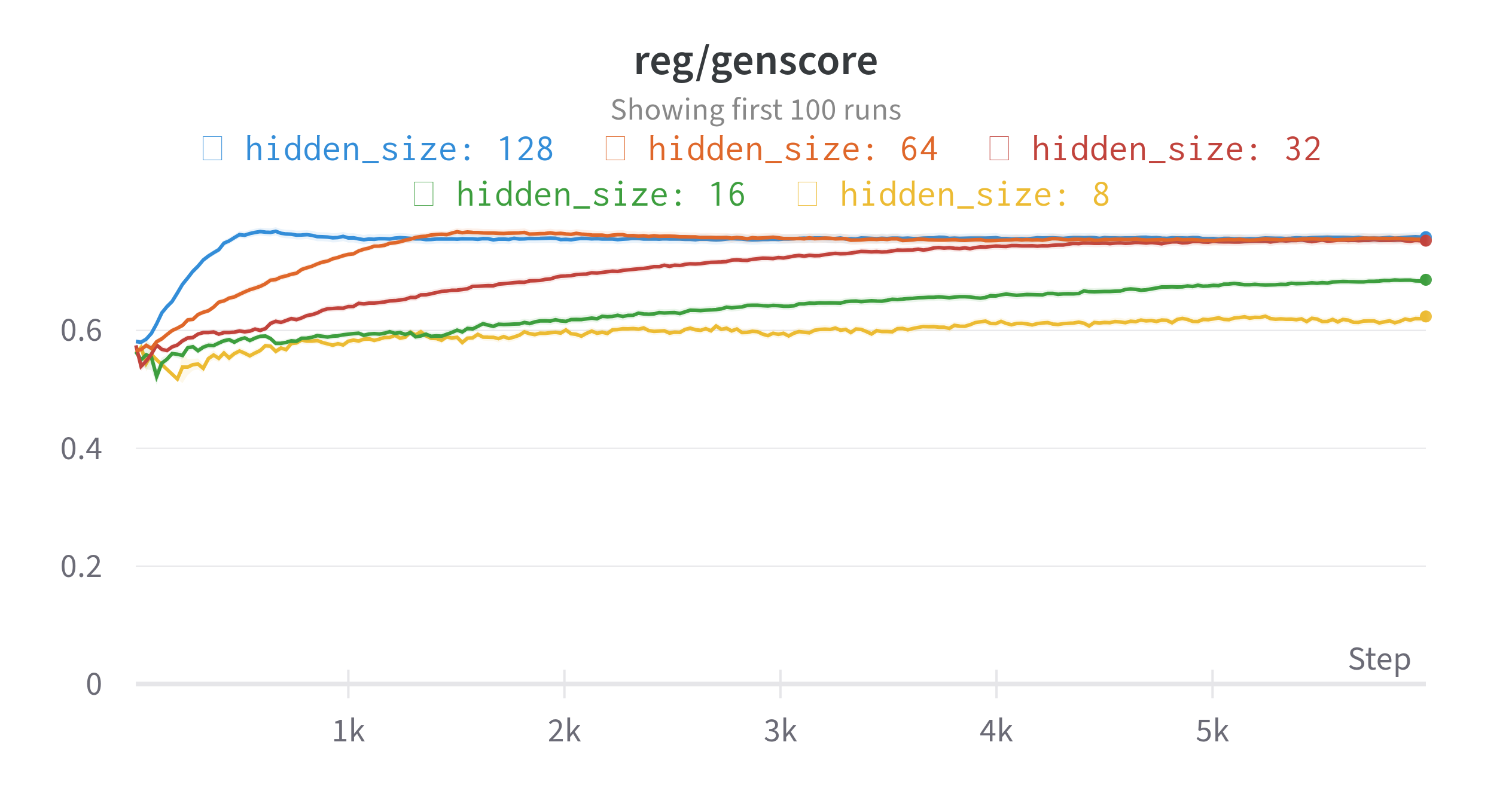}
    \caption{Average generalization systematicity across input languages as a function of the size of the neural networks' hidden layers}
    \label{fig:hiddensize:reg-genscore}
\end{figure}

We vary the hidden size and plot the average scores over the 10 input languages.
Figure~\ref{fig:hiddensize:mem-prodsim} shows the production similarity during memorization.
Figure~\ref{fig:hiddensize:reg-prodsim} shows the production similarity between neural agents and human learners during testing.
Figure~\ref{fig:hiddensize:reg-genscore} shows the generalization score of neural agents.

\subsection*{Sensitivity to the Scaling Factor for the Contrastive Loss Term}\label{sub:conloss}

\begin{figure}[p]
    \centering
    \includegraphics[width=\scalefig\textwidth]{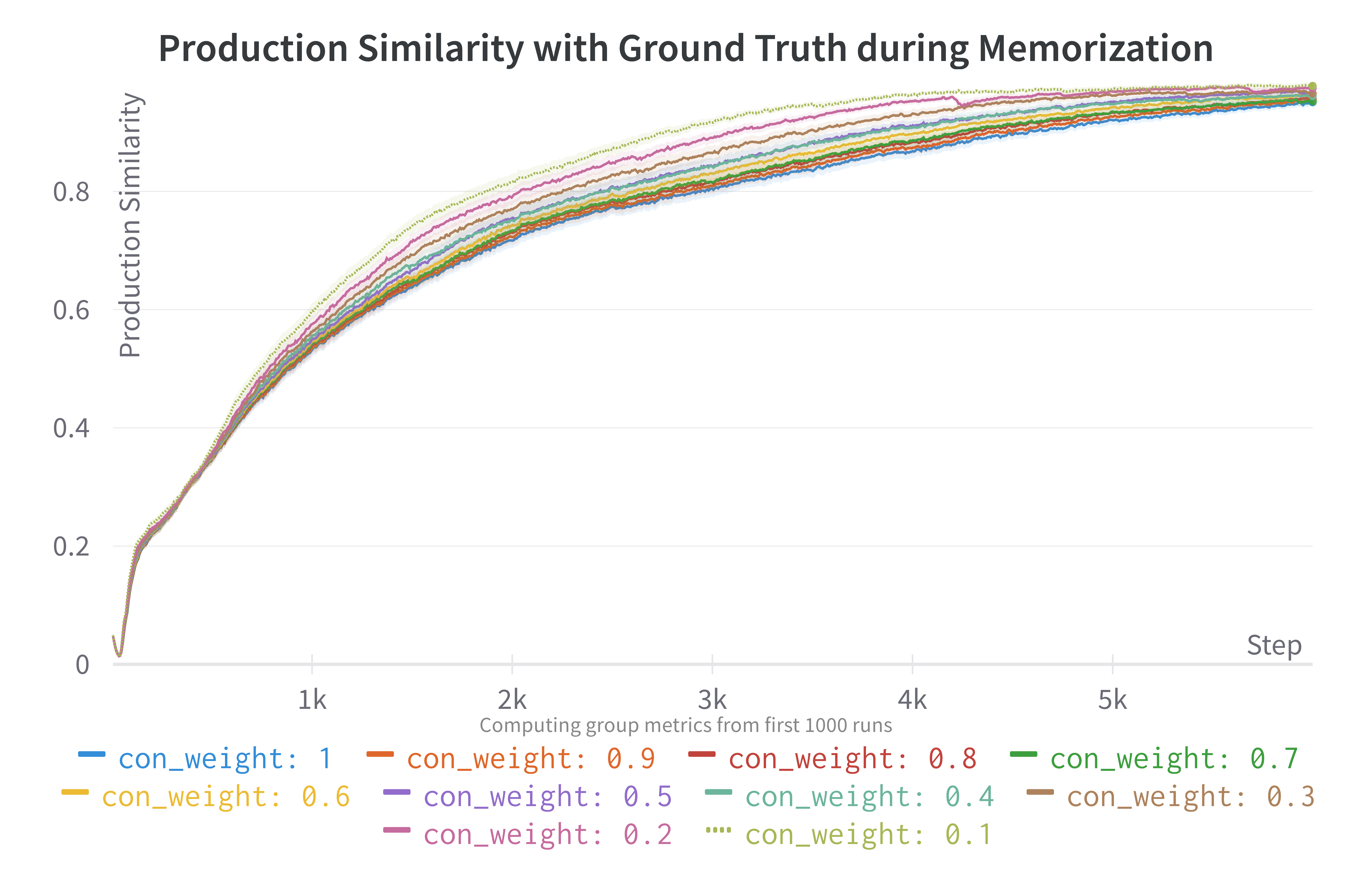}
    \caption{Average production similarity to ground truth across input languages during memorization as a function of the scaling factor for the contrastive loss term.}
    \label{fig:conloss:mem-prodsim}
\end{figure}

\begin{figure}[p]
    \centering
    \includegraphics[width=\scalefig\textwidth]{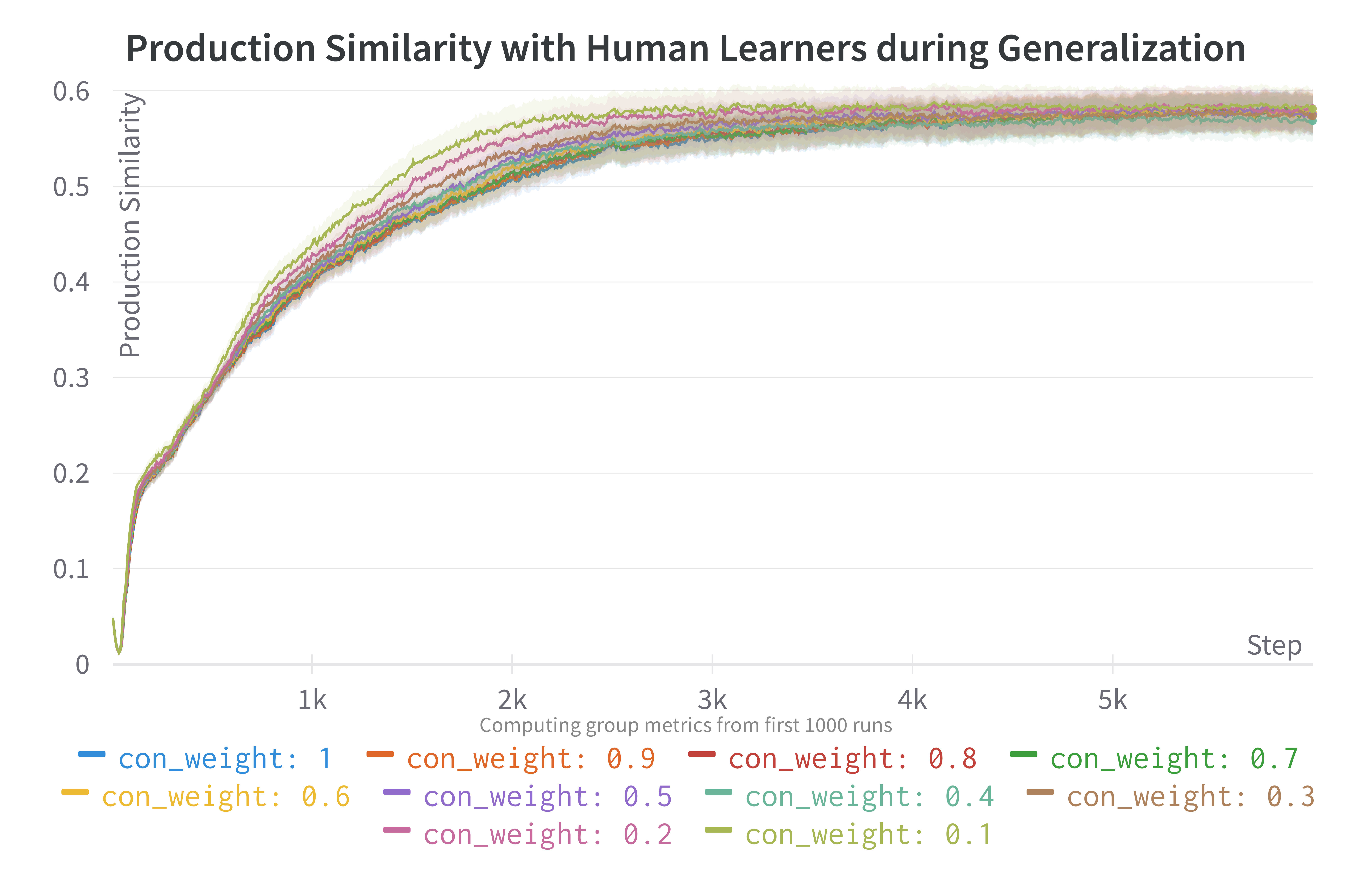}
    \caption{Average production similarity with human learners across input languages during generalization as a function of the scaling factor for the contrastive loss term.}
    \label{fig:conloss:reg-prodsim}
\end{figure}

\begin{figure}[p]
    \centering
    \includegraphics[width=\scalefig\textwidth]{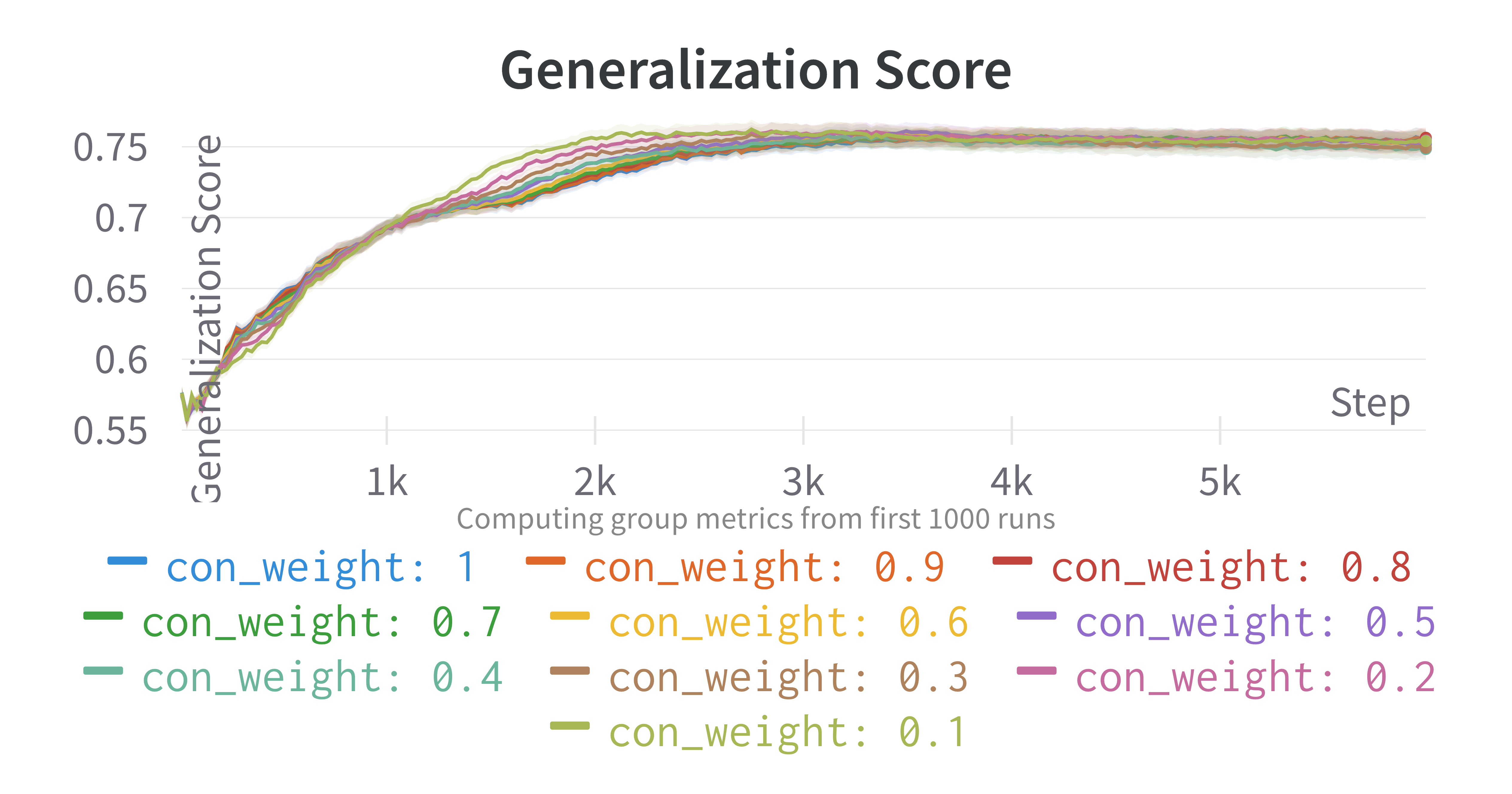}
    \caption{Average generalization score across input languages as a function of the scaling factor for the contrastive loss term.}
    \label{fig:conloss:reg-genscore}
\end{figure}

We experiment with different scaling factors $\alpha$ for the contrastive loss term.
Figure~\ref{fig:conloss:mem-prodsim} shows the results for production similarity with ground truth during the memorization test.
Here, we report the average across all input languages with different degrees of structuredness.
Figure~\ref{fig:conloss:reg-prodsim} shows the results for production similarity with human learners during the generalization test. Again, a scaling factor of $0.1$ leads to the best results in terms of learning speed. However, there is little difference to using a scaling factor of $0.2$.
Figure~\ref{fig:conloss:reg-genscore} shows the results for generalization score (scaled to $[0,1]$). We observe that the scaling factor of $0.1$ has advantages in learning speed. Starting at step 1,300, the generalization score increases faster with scaling factor $0.1$ than with other scaling factors.

\subsection*{Structure effect does not depend on specific hyperparameter choices}
\begin{figure}[htp!]
\centering
\includegraphics[width=\textwidth]{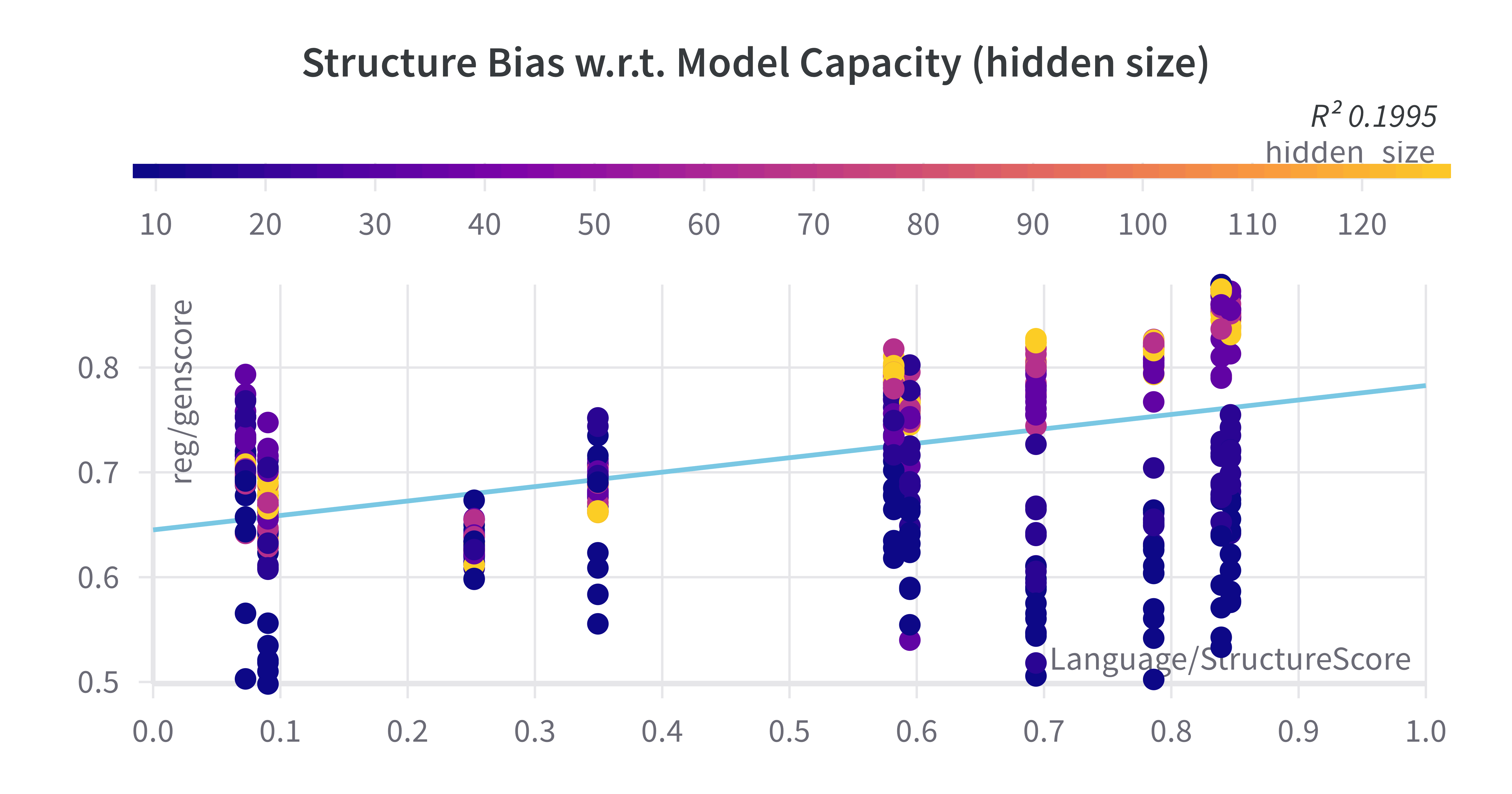}
\caption{Relationship between compositional structure and generalization score with different values for hidden size controlling the model capacity}\label{fig:struct-bias-wrt-hidden-size}
\end{figure}

\begin{figure}[htp!]
\centering
\includegraphics[width=\textwidth]{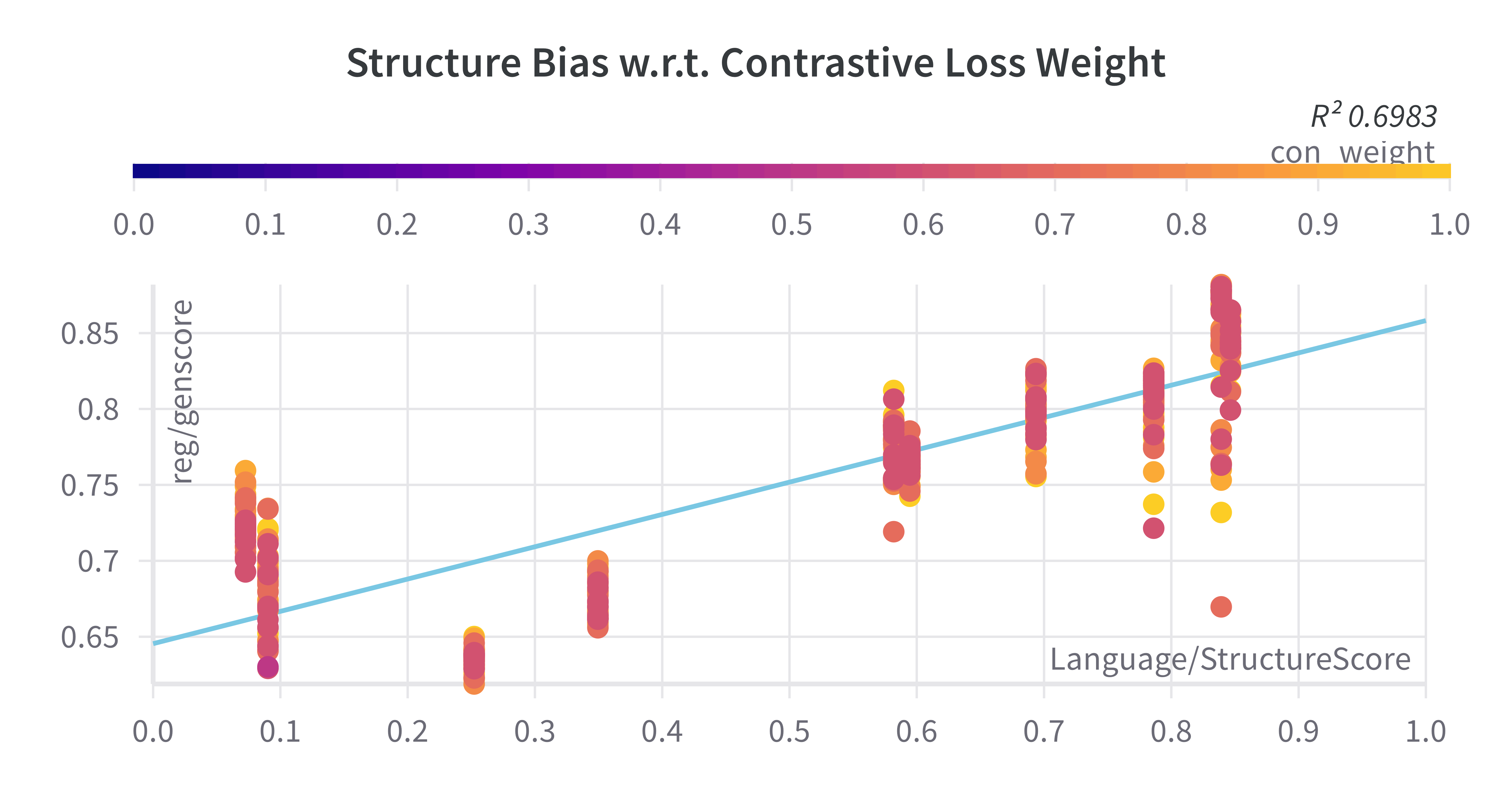}
\caption{Relationship between compositional structure and generalization score with different values for contrastive loss weight controlling the influence of the contrastive objective used for guessing tasks}\label{fig:struct-bias-wrt-con-loss}
\end{figure}

Figure~\ref{fig:struct-bias-wrt-hidden-size} and
Figure~\ref{fig:struct-bias-wrt-con-loss} show the relationship between the
degree of compositional structure and the generalizations core with respect to
the hidden size and the scaling factor for the contrastive loss.

\newpage
\section*{Example Data}

We provide example data from the memorization test in
Table~\ref{tab:mem-examples} and from the generalization test in
Table~\ref{tab:reg-examples}. The sample is stratified with respect to the
by-producer average production similarity to human participants. For each
percentile out of 0, 25, 50, 75, and 100, we randomly sample 8 items. All
samples are taken at the end of training (after epoch 100).

\begin{table}[p]
    \centering
    \caption{Example Data from the Memorization Test.}\label{tab:mem-examples}
    \begin{tabular}{rrlrrlll}
\toprule
Sim. to Humans & Producer & Lang. & Shape & Angle & True Label & Participant Label & RNN Label \\
\midrule
0.247826 & 4082 & S1 & 4 & 225 & wufk & moof & wufk \\
0.247826 & 4082 & S1 & 2 & 330 & weft & puif & weft \\
0.247826 & 4082 & S1 & 1 & 240 & smoogg & huift & smoogg \\
0.247826 & 4082 & S1 & 4 & 300 & suith & keft & suith \\
0.247826 & 4082 & S1 & 3 & 135 & spuut & weft & spuut \\
0.247826 & 4082 & S1 & 1 & 60 & koof & puif & koof \\
0.247826 & 4082 & S1 & 4 & 120 & wafk & sgeft & wafk \\
0.247826 & 4082 & S1 & 3 & 270 & huipt & koom & huipt \\
0.700932 & 1093 & B2 & 4 & 270 & gnt & gnt & gnt \\
0.700932 & 1093 & B2 & 4 & 135 & gnts & gntsi & gnts \\
0.700932 & 1093 & B2 & 2 & 330 & gmps & gmps & gmps \\
0.700932 & 1093 & B2 & 2 & 120 & wangsus & wangsus & wangsu \\
0.700932 & 1093 & B2 & 2 & 45 & gempt & wangsus & gempt \\
0.700932 & 1093 & B2 & 4 & 330 & gntsi & gnit & gntsi \\
0.700932 & 1093 & B2 & 2 & 300 & wangsu & gmps & wangsu \\
0.700932 & 1093 & B2 & 1 & 30 & sket & sket & sket \\
0.883282 & 3090 & S5 & 2 & 180 & nif-a & nif-a & nif-as \\
0.883282 & 3090 & S5 & 1 & 90 & wef-t & wef-t & wef-t \\
0.883282 & 3090 & S5 & 1 & 150 & wef-aat & wef-att & wef-aat \\
0.883282 & 3090 & S5 & 1 & 240 & wef-ssa & wef-aas & wef-ssa \\
0.883282 & 3090 & S5 & 4 & 210 & smut-aas & smut-aas & smut-aas \\
0.883282 & 3090 & S5 & 3 & 180 & pti-a & pti-a & pti-a \\
0.883282 & 3090 & S5 & 3 & 315 & pti-kks & pti-ssk & pti-kks \\
0.883282 & 3090 & S5 & 2 & 225 & nif-as & nif-as & nif-as \\
0.956522 & 1020 & B1 & 2 & 330 & wak-ta & wak-ta & wak-ta \\
0.956522 & 1020 & B1 & 4 & 225 & gtgt & gtgt & gtgt \\
0.956522 & 1020 & B1 & 4 & 120 & ftft & ftft & ftft \\
0.956522 & 1020 & B1 & 2 & 150 & hehi & hehi & hehi \\
0.956522 & 1020 & B1 & 1 & 210 & ha-ia & ha-ia & ha-ia \\
0.956522 & 1020 & B1 & 3 & 45 & fiti & fiti & fiti \\
0.956522 & 1020 & B1 & 4 & 60 & kite & fik & kite \\
0.956522 & 1020 & B1 & 4 & 360 & pepepe & pepepe & pepepe \\
1.000000 & 1008 & B4 & 4 & 225 & fak-huif & fak-huif & fak-huif \\
1.000000 & 1008 & B4 & 3 & 45 & muif-a & muif-a & muif-a \\
1.000000 & 1008 & B4 & 4 & 120 & fak-e & fak-e & fak-e \\
1.000000 & 1008 & B4 & 1 & 315 & fas-pok & fas-pok & fas-pok \\
1.000000 & 1008 & B4 & 3 & 135 & muif-e & muif-e & muif-e \\
1.000000 & 1008 & B4 & 2 & 135 & pok-e & pok-e & pok-e \\
1.000000 & 1008 & B4 & 4 & 300 & fak-pok & fak-pok & fak-pok \\
1.000000 & 1008 & B4 & 2 & 210 & pok-huif & pok-huif & pok-huif \\
\bottomrule
\end{tabular}

\end{table}

\begin{table}[p]
    \centering
    \caption{Example Data from the Generalization Test.}\label{app:tab:reg-examples}
    \begin{tabular}{rrlrrlll}
\toprule
Sim. to Humans & Producer & Lang. & Shape & Angle & Participant Label & RNN Label \\
\midrule
0.092308 & 1048 & B1 & 2 & 360 & kokoke & seefe \\
0.092308 & 1048 & B1 & 1 & 225 & po-ti & ha-ia \\
0.092308 & 1048 & B1 & 3 & 90 & ghio & mimi \\
0.092308 & 1048 & B1 & 4 & 240 & khio & gtgt \\
0.092308 & 1048 & B1 & 3 & 150 & ptpt & mimi \\
0.092308 & 1048 & B1 & 2 & 300 & ko-toe & wak-ta \\
0.092308 & 1048 & B1 & 4 & 210 & ka-ia & gtgt \\
0.092308 & 1048 & B1 & 2 & 225 & haia & pooti \\
0.443223 & 8064 & B2 & 3 & 150 & wangsi & wangsuu \\
0.443223 & 8064 & B2 & 4 & 225 & gntsoe & gntuu \\
0.443223 & 8064 & B2 & 1 & 135 & sketsi & gesh \\
0.443223 & 8064 & B2 & 4 & 360 & gnt & skek \\
0.443223 & 8064 & B2 & 4 & 60 & gmpsi & skek \\
0.443223 & 8064 & B2 & 2 & 270 & wng & wangsuu \\
0.443223 & 8064 & B2 & 2 & 360 & wang & gempt \\
0.443223 & 8064 & B2 & 3 & 225 & wangsuus & wangsoe \\
0.590884 & 1083 & S3 & 4 & 60 & fuottee & fuoto-o-o-o \\
0.590884 & 1083 & S3 & 4 & 150 & fuottoo & fuottii \\
0.590884 & 1083 & S3 & 1 & 30 & fewo-o-o-o & fewen \\
0.590884 & 1083 & S3 & 3 & 60 & powi & powu-u-u \\
0.590884 & 1083 & S3 & 1 & 225 & fewo-o-o-o & fewo-o-o \\
0.590884 & 1083 & S3 & 3 & 225 & powee & powwoo \\
0.590884 & 1083 & S3 & 2 & 360 & asken & asko-o-o \\
0.590884 & 1083 & S3 & 4 & 330 & fuottoa & fuotio \\
0.772497 & 2058 & B4 & 2 & 360 & pok-i & pok \\
0.772497 & 2058 & B4 & 4 & 330 & fak-pok & fas-i \\
0.772497 & 2058 & B4 & 4 & 60 & fak-a & fak-e \\
0.772497 & 2058 & B4 & 1 & 30 & fas-a & fas-a \\
0.772497 & 2058 & B4 & 2 & 300 & pok & pok-u \\
0.772497 & 2058 & B4 & 4 & 90 & fak-u & fak-e \\
0.772497 & 2058 & B4 & 4 & 150 & fak-w-w-e & fak-e \\
0.772497 & 2058 & B4 & 3 & 225 & muif-huif & muif-huif \\
0.923443 & 5016 & B4 & 4 & 90 & fak-iii & fak-e \\
0.923443 & 5016 & B4 & 1 & 120 & fas-e & fas-e \\
0.923443 & 5016 & B4 & 2 & 300 & pok & pok \\
0.923443 & 5016 & B4 & 3 & 60 & muif-a & muif-a \\
0.923443 & 5016 & B4 & 3 & 360 & muif-i & muif-a \\
0.923443 & 5016 & B4 & 4 & 150 & fak-e & fak-e \\
0.923443 & 5016 & B4 & 3 & 225 & muif-huif & muif-huif \\
0.923443 & 5016 & B4 & 4 & 150 & fak-e & fak-e \\
\bottomrule
\end{tabular}

\end{table}

\end{document}